\documentclass{article}

\usepackage{arxiv}

\usepackage[T1]{fontenc}
\usepackage[utf8]{inputenc}

\usepackage[protrusion=true, expansion=true]{microtype}
\DisableLigatures{encoding = *, family = *}

\usepackage[table]{xcolor}
\usepackage{array, colortbl, tabularx, booktabs, multirow, longtable}
\usepackage{tcolorbox}
\definecolor{headergray}{rgb}{0.2, 0.2, 0.2}
\definecolor{catgreen}{rgb}{0.5, 0.75, 0.2}
\definecolor{modelblue}{rgb}{0.2, 0.7, 0.9}
\definecolor{white}{rgb}{1,1,1}

\usepackage{amsmath, amsfonts}
\usepackage[linesnumbered,ruled,vlined]{algorithm2e}

\usepackage{graphicx}
\graphicspath{{./images/}}
\usepackage{tikz}
\usetikzlibrary{shapes.geometric, arrows, trees, positioning, plotmarks, calc, fit, backgrounds}

\usepackage{float}
\usepackage{adjustbox}
\usepackage{listings}
\usepackage{caption}
\usepackage{lipsum} % Placeholder text
\usepackage[edges]{forest}

\usepackage[hyphens]{url}
\usepackage[colorlinks=true, urlcolor=blue, linkcolor=blue, citecolor=blue]{hyperref}
\usepackage{etoolbox}
\apptocmd{\UrlBreaks}{\do\/\do\-}{}{}

\title{Explaining Large Language Models with gSMILE}

\author{
  Zeinab Dehghani \\
  University of Hull \\
  United Kingdom \\
  \texttt{z.dehghani-2023@hull.ac.uk}
  \and
  Mohammed Naveed Akram \\
  Fraunhofer IESE \\
  Germany \\
  \texttt{naveed.akram@iese.fraunhofer.de}
  \and
  Koorosh Aslansefat \\
  University of Hull \\
  United Kingdom \\
  \texttt{k.aslansefat@hull.ac.uk}
  \and
  Adil Khan \\
  University of Hull \\
  United Kingdom \\
  \texttt{a.m.khan@hull.ac.uk}
  \and
  Yiannis Papadopoulos \\
  University of Hull \\
  United Kingdom \\
  \texttt{y.i.papadopoulos@hull.ac.uk}
}

\begin{document}

\maketitle

\begin{abstract}

Large Language Models (LLMs) such as GPT, LLaMA, and Claude achieve remarkable performance in text generation but remain opaque in their decision-making processes, limiting trust and accountability in high-stakes applications. We present gSMILE (generative SMILE), a model-agnostic, perturbation-based framework for token-level interpretability in LLMs. Extending the SMILE methodology, gSMILE utilises controlled prompt perturbations, Wasserstein distance metrics, and weighted linear surrogates to identify input tokens that have the most significant impact on the output. This process enables the generation of intuitive heatmaps that visually highlight influential tokens and reasoning paths. We evaluate gSMILE across leading LLMs (OpenAI’s gpt-3.5-turbo-instruct, LLaMA 3.1 Instruct Turbo, and Anthropic’s Claude 2.1) using attribution metrics such as fidelity, consistency, stability, faithfulness, and accuracy. Results show that gSMILE delivers reliable human-aligned attributions, with Claude 2.1 excelling in attention fidelity and GPT-3.5 achieving the highest output consistency. These findings demonstrate gSMILE's ability to strike a balance between model performance and interpretability, enabling more transparent and trustworthy AI systems.

\end{abstract}

\begin{figure}[H]
    \centering
    \includegraphics[width=0.8\textwidth]{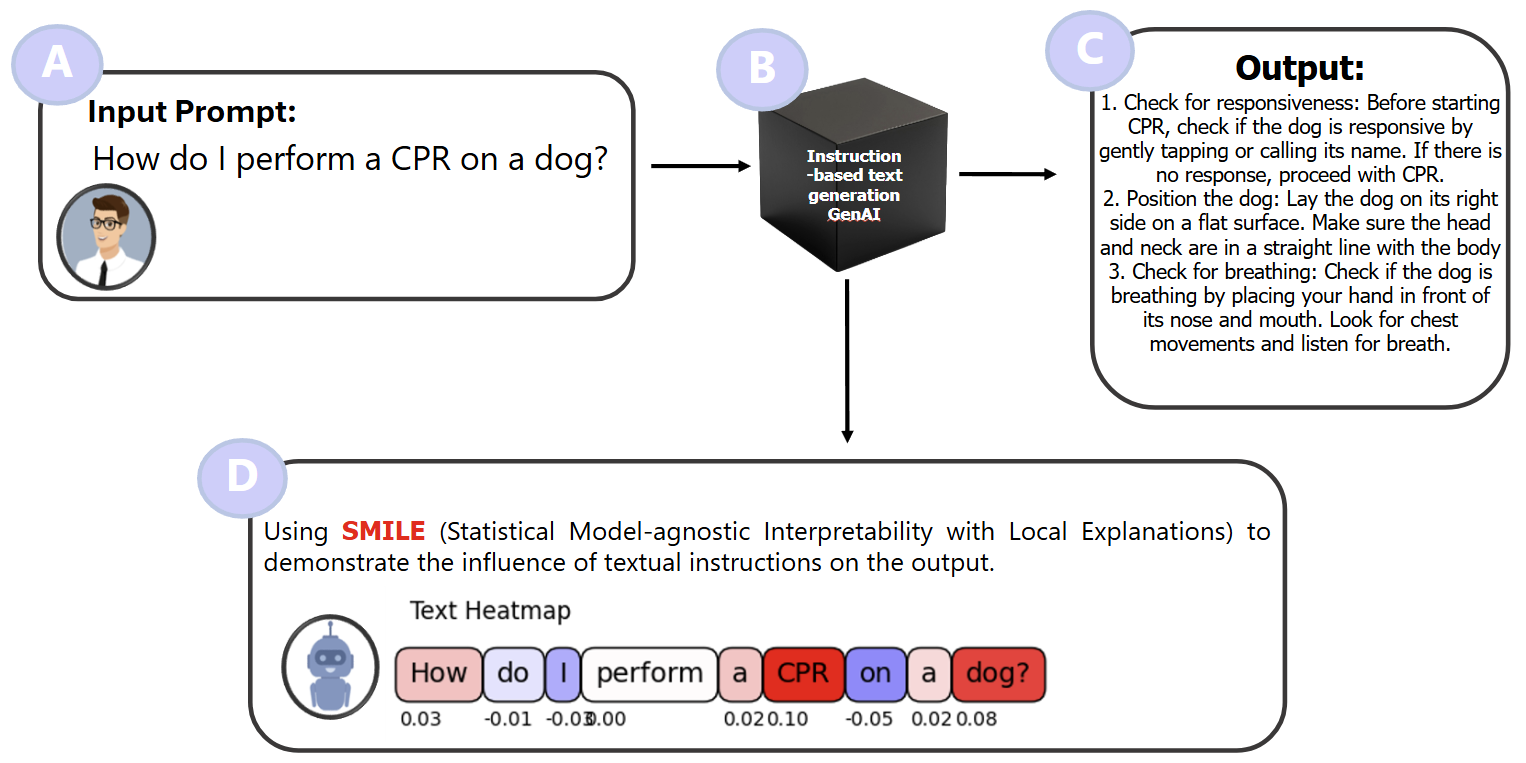}
    \caption{Overview of gSMILE interpretability process: An input prompt is processed by a transformer-based LLM, and gSMILE generates a token-level heatmap highlighting influential words contributing to the output.}
    \label{fig:totalflow}
\end{figure}

\section{Introduction}

Large Language Models (LLMs) such as GPT‑4~\cite{achiam2023gpt}, Claude~\cite{anthropic2023claude}, and LLaMA~\cite{touvron2023llama} have rapidly become central to artificial intelligence, powering applications in healthcare, education, law, and scientific research~\cite{aljohani2025comprehensive,ghosh2024logical,deng2024information,rane2023contribution}. Built on the Transformer architecture~\cite{vaswani2017attention} and trained on massive corpora, they generate fluent, context‑aware responses for tasks ranging from dialogue and summarisation to reasoning and information extraction~\cite{deng2024information}.  

Today, LLMs are embedded in widely used technologies, assisting clinicians in healthcare~\cite{aljohani2025comprehensive}, supporting personalised education~\cite{ghosh2024logical}, streamlining customer service~\cite{deng2024information}, and aiding legal research~\cite{rane2023contribution}. Their growing influence underscores both their transformative potential and the urgent need to address concerns about reliability, transparency, and trust.

Despite their remarkable capabilities, a fundamental question remains: how much control do we truly have over LLM behaviour, and to what extent do we understand their decision-making processes? While they can generate compelling outputs, studies show that these responses are not always reliable, often containing hallucinations or logical inconsistencies that damage trust~\cite{aljohani2025comprehensive,ghosh2024logical}. In practice, users are left uncertain about why a model provides one answer instead of another, raising concerns about whether we genuinely comprehend the mechanisms driving their outputs~\cite{xu2024llms}.

Recent work has also highlighted the critical role of prompt design in shaping LLM performance. Simple changes to instructions, for example, adding phrases such as ``Let’s think step by step'' or ``Take a deep breath and work on this problem step-by-step, have been shown to improve accuracy on reasoning benchmarks significantly. OPRO reports that the latter phrase boosts GSM8K accuracy to 80.2\%, compared to 71.8\% with the baseline step-by-step prompt~\cite{yang2023large}. Similarly, Self-Supervised Prompt Optimisation (SPO) demonstrates that optimised prompts, even without external references, consistently improve results across multiple datasets~\cite{xiang2025self}. These findings underline the power of prompt engineering: if we understood more clearly which features of a prompt drive performance, we could design instructions that yield more reliable outputs. Interpretability methods can provide the insights needed to guide such prompt design.

Moreover, interpretability approaches must be model‑agnostic. While Anthropic has recently showcased advanced interpretability research for Claude 3.5, using techniques such as circuit tracing and attribution graphs to visualise how the model plans, manipulates concepts, and sometimes even fabricates reasoning~\cite{anthropic2025attribution}, such access to internal mechanisms is limited to others with privileged model access. For most practitioners and end users, leading LLMs like GPT and Claude are available only through APIs that restrict visibility into their parameters and architectures. As a result, practical interpretability techniques must be designed to explain model behaviour without relying on direct access to internal weights or structural details~\cite{deng2024information}.

To this end, we propose \textbf{gSMILE}, a generalised and LLM-oriented extension of the SMILE (Statistical Model-Agnostic Interpretability with Local Explanations) framework~\cite{aslansefat2023explaining}. gSMILE interprets model outputs by perturbing input prompts, measuring output shifts using Wasserstein distance, and fitting a local linear surrogate model. The result is a weighted heatmap that visualises how individual words in a prompt contribute to the model's response.

We evaluate gSMILE across multiple large language models (LLMs) and prompt types using quantitative metrics, including ATT fidelity, ATT consistency, ATT stability, ATT faithfulness and ATT accuracy. Our findings demonstrate that gSMILE yields stable, interpretable, and reliable attributions, rendering it suitable for practical applications in real-world scenarios involving instruction-based generation.

\textbf{The key contributions of this paper are:}

\begin{itemize}
    \item \textbf{Token-level interpretability for LLMs:} We present gSMILE, a model-agnostic method that quantifies the influence of each input token on the output of large language models using perturbation-based analysis and Wasserstein distance.
    
    \item \textbf{Visual attribution through weighted heatmaps:} The framework generates intuitive heatmaps that highlight high-impact tokens, enabling human-aligned interpretation of model behaviour.
    
    \item \textbf{Robust cross-model evaluation:}We evaluate gSMILE on multiple instruction-tuned LLMs using diverse metrics, including ATT accuracy, ATT fidelity, ATT stability, ATT faithfulness and ATT consistency, that demonstrate its reliability and adaptability.
\end{itemize}

\section*{List of Abbreviations}
\begin{table}[H]
\centering
\renewcommand{\arraystretch}{1.2}
\begin{tabular}{ll}
\hline
\textbf{Abbreviation} & \textbf{Definition} \\
\hline
LLM & Large Language Model \\
XAI & Explainable Artificial Intelligence \\
SMILE & Statistical Model-agnostic Interpretability with Local Explanations \\
gSMILE & Generative SMILE (proposed method) \\
WMD & Word Mover's Distance \\
IWMD & Inner Word Mover’s Distance (input-level distance) \\
OWMD & Outcome Word Mover’s Distance (output-level distance) \\
WD & Wasserstein Distance \\
ATT & Attribution \\
TV & Total Variation distance \\
P & Norm order in Wasserstein-$P$ metric \\
$ \pi^{(n)} $ & Black-box text generation model \\
$ \Delta(x, \hat{x}_j) $ & Output-level semantic shift \\
$ w_j $ & Similarity-based kernel weight \\
$ z_j $ & Feature vector representation of perturbed prompt \\
\hline
\end{tabular}
\end{table}

\section{Literature review}
\label{sec:headings}

This section provides an overview of recent advancements in large language models (LLMs) and their increasing impact across various applications. We also introduce Explainable Artificial Intelligence (XAI), which aims to make AI models more transparent, interpretable, and trustworthy. As LLMs grow in complexity and usage, the need to understand their decision-making processes becomes more critical. Therefore, we highlight the emerging efforts to develop explainable large language models (LLMs), outlining key challenges and recent techniques that aim to improve their interpretability and foster greater trust in their outputs.

\subsection{Large Language Models}

Large language models (LLMs) have rapidly advanced due to progress in deep learning~\cite{lecun2015deep}, the availability of massive datasets~\cite{zaharia2016apache}, and increasingly powerful hardware. Trained on large text corpora, they can generate fluent responses for tasks such as question answering, summarisation, translation, and content creation. Key innovations include new architectures~\cite{vaswani2017attention}, training strategies~\cite{raffel2020exploring}, and alignment methods~\cite{ouyang2022training}.

Several models have shaped the field. OpenAI’s GPT series~\cite{brown2020language} popularised autoregressive transformers through ChatGPT. Google’s T5~\cite{raffel2020exploring} reframed NLP as a text-to-text problem, while PaLM~\cite{chowdhery2023palm} scaled multitask learning. Anthropic’s Claude~\cite{anthropic2025attribution} introduced constitutional AI to emphasise safety, and Google’s Gemma~\cite{team2024gemma} advanced reasoning-focused applications. Meta’s LLaMA models~\cite{touvron2023llama} provided a scalable open-source alternative.

Community-driven projects also played a significant role. EleutherAI’s GPT-NeoX~\cite{black2022gpt} offered an open substitute for proprietary GPT models. AI21’s Jurassic-2~\cite{lieber2021jurassic} targeted enterprise integration, and Cohere’s Command R+~\cite{gao2023retrieval} combined generation with retrieval. Earlier, BERT~\cite{devlin2019bert} transformed bidirectional understanding. More recent lightweight models such as Phi-4, Mistral, and Alibaba’s Qwen 2.5 focus on efficiency, multilinguality, and adaptability.

The latest generation continues this trend. DeepSeek-LLM~\cite{bi2024deepseek}, Mixtral~\cite{jiang2024mixtral}, and StableLM prioritise open and efficient deployment, while DeepSeek-R1 highlights reasoning. Others push specialisation: Qwen2.5-Max uses Mixture-of-Experts for scalability, OpenAI’s o3-mini is designed for low-latency use, and DeepSeek-V3 has reached GPT-4-level performance in an open-source setting.

As shown in Fig.~\ref{fig:llm-full-table}, today’s LLMs are not only powerful text generators but also increasingly efficient, transparent, and aligned with human values. They continue to shape the state of the art in both research and industry.

\begin{figure}[H]
\centering
\begin{tikzpicture}[
    font=\sffamily\small,
    box/.style={minimum height=1.4cm, align=center, text width=4.5cm, draw=black, fill=white, text=black, rounded corners=6pt},
    cat/.style={box, fill=blue!5},
    model/.style={box},
    desc/.style={box},
    header/.style={box, fill=blue!20, font=\bfseries}
]
% Header Row
\node[header] (h1) at (0,0) {Category};
\node[header] (h2) at (5,0) {Models};
\node[header] (h3) at (10,0) {Description};

% General-Purpose
\node[cat]   at (0,-1.8) {General-Purpose};
\node[model] at (5,-1.8) {GPT-4~\cite{achiam2023gpt}, GPT-4o~\cite{hurst2024gpt},\\ Claude 3~\cite{anthropic2024claude3}, Claude 2.1~\cite{claude2024} Gemini 1.5~\cite{team2024gemini},\\ Command R+~\cite{gao2023retrieval}, Mistral~\cite{jiang2023mistral7b}};
\node[desc]  at (10,-1.8) {Versatile for wide\\ range of NLP tasks};

% Open-Source
\node[cat]   at (0,-3.6) {Open-Source};
\node[model] at (5,-3.6) {LLaMA 3~\cite{grattafiori2024llama}, Mistral~\cite{jiang2023mistral7b},\\ Mixtral~\cite{jiang2024mixtral}, Falcon 180B~\cite{almazrouei2023falcon},\\ DeepSeek-V3~\cite{liu2024deepseek}, OpenChat~\cite{wang2023openchat}};
\node[desc]  at (10,-3.6) {Customizable models for\\ research applications};

% Retrieval-Augmented
\node[cat]   at (0,-5.4) {Retrieval-Augmented};
\node[model] at (5,-5.4) {Command R+~\cite{gao2023retrieval},\\ GPT-4 Bing~\cite{achiam2023gpt}, Claude 3 RAG~\cite{anthropic2024claude3},\\ Gemini Search~\cite{team2024gemini}};
\node[desc]  at (10,-5.4) {Enhance information\\ retrieval accuracy,\\ reduce misinformation};

% Long-Text Processing
\node[cat]   at (0,-7.2) {Long-Text Processing};
\node[model] at (5,-7.2) {Claude 3~\cite{anthropic2024claude3}, Gemini 1.5~\cite{team2024gemini},\\ GPT-4o~\cite{hurst2024gpt}, Grok-1.5~\cite{xai2024grok}};
\node[desc]  at (10,-7.2) {Specialized in\\ processing long-form text};

% Advanced Reasoning
\node[cat]   at (0,-9) {Advanced Reasoning};
\node[model] at (5,-9) {GPT-4~\cite{achiam2023gpt}, Claude 3 Opus~\cite{anthropic2024claude3},\\ Gemini 1.5~\cite{team2024gemini}, Mixtral~\cite{jiang2024mixtral},\\ Grok-1.5~\cite{xai2024grok}};
\node[desc]  at (10,-9) {Logical reasoning,\\ multi-task learning,\\ problem-solving};

% Ethically-Aligned
\node[cat]   at (0,-10.8) {Safety-Focused Design};
\node[model] at (5,-10.8) {Claude 3~\cite{anthropic2024claude3}, Gemini~\cite{team2024gemini},\\ GPT-4~\cite{achiam2023gpt}, Grok~\cite{xai2024grok}};
\node[desc]  at (10,-10.8) {Emphasize alignment\\ with safety rules and\\ bias mitigation};

% MoE
\node[cat]   at (0,-12.6) {Mixture-of-Experts (MoE)};
\node[model] at (5,-12.6) {Mixtral~\cite{jiang2024mixtral}, Qwen2.5-Max~\cite{hui2024qwen2},\\ GShard~\cite{lepikhin2020gshard}, Switch Transformer~\cite{fedus2022switch}};
\node[desc]  at (10,-12.6) {Improved efficiency\\ and performance via MoE};

% Multimodal
\node[cat]   at (0,-14.4) {Multimodal};
\node[model] at (5,-14.4) {GPT-4o~\cite{hurst2024gpt}, Gemini 1.5~\cite{team2024gemini},\\ Claude 3~\cite{anthropic2024claude3}, Kosmos-2~\cite{peng2023kosmos},\\ Grok-1.5V~\cite{xai2024grok}};
\node[desc]  at (10,-14.4) {Capable of processing\\ both text and visual inputs};

% Multilingual
\node[cat]   at (0,-16.2) {Multilingual};
\node[model] at (5,-16.2) {BLOOM~\cite{le2023bloom}, XGLM~\cite{chi2021xlm},\\ mBERT~\cite{devlin2019bert}, XLM-R~\cite{conneau2019unsupervised},\\ ByT5~\cite{xue2022byt5}};
\node[desc]  at (10,-16.2) {Optimized for cross-lingual\\ understanding and translation};

% Domain-Specific
\node[cat]   at (0,-18) {Domain-Specific};
\node[model] at (5,-18) {Med-PaLM 2~\cite{singhal2025toward}, Galactica~\cite{taylor2023galactica},\\ Legal-BERT~\cite{chalkidis2020legal}, StarCoder2~\cite{lozhkov2024starcoder},\\ Code LLaMA~\cite{roziere2023code}};
\node[desc]  at (10,-18) {Tailored for specific tasks\\ such as medical, legal, or code};

% Instruction-Tuned
\node[cat]   at (0,-19.8) {Instruction-Tuned};
\node[model] at (5,-20) {Alpaca~\cite{taori2023alpaca}, Vicuna~\cite{chiang2023vicuna},\\ ChatGLM~\cite{du2021glm}, Baize~\cite{xu2023baize},\\ OpenChat~\cite{wang2023openchat}, GPT-3.5~\cite{ye2023comprehensive}, LLaMA3.1~\cite{llama2024huggingface}};
\node[desc]  at (10,-19.8) {Tuned for better instruction-following\\ and chat capabilities};

% Lightweight
\node[cat]   at (0,-21.6) {Lightweight / Edge};
\node[model] at (5,-21.6) {Phi-2~\cite{li2023textbooks}, TinyLLaMA~\cite{zhang2024tinyllama},\\ DistilGPT2~\cite{sanh2019distilbert}, GPT-2~\cite{radford2019gpt2},\\ LLaMA 2 7B~\cite{touvron2023llama}};
\node[desc]  at (10,-21.6) {Designed for fast inference\\ and deployment on edge devices};

\end{tikzpicture}
\caption{Extended LLM Categories, Models, and Descriptions}
\label{fig:llm-full-table}
\end{figure}

\subsection{Explainable AI (XAI)}
The increasing complexity of machine learning models has made understanding how they make decisions more critical than ever. Ensuring transparency, fairness, and reliability is crucial, so explainability has become a key focus in AI research~\cite{adadi2018peeking, doshi2017towards}.

Explainability methods can be broadly categorised as intrinsic or post-hoc. Intrinsic methods involve inherently interpretable models due to their simplicity, such as linear regression or decision trees~\cite{molnar2020interpretable}. These models are often preferred in applications where transparency is critical, but they may lack the predictive power of more complex models~\cite{rudin2019stop}.

In contrast, post-hoc methods, such as Local Interpretable Model-agnostic Explanations (LIME)~\cite{ribeiro2016should} and SHAP (Shapley Additive Explanations)~\cite{lundberg2017unified}, are applied after model training to explain the decisions of complex, often ``black-box'' models like neural networks and ensemble methods. These approaches provide valuable insights without altering the underlying model structure~\cite{guidotti2018survey}. Post-hoc explainability methods can be classified into global and local explanations.

Global explanations offer an overarching view of the model's decision-making patterns across an entire dataset, facilitating the identification of trends and biases~\cite{molnar2020interpretable, du2019techniques}. In contrast, local explanations focus on specific predictions, providing a detailed breakdown of how individual inputs influence model outputs. LIME and its numerous enhancements fall under local explainability methods and are designed to provide granular, case-specific interpretations~\cite{ribeiro2016should}.

\begin{figure}[H]
    \centering
    \includegraphics[width=0.8\textwidth]{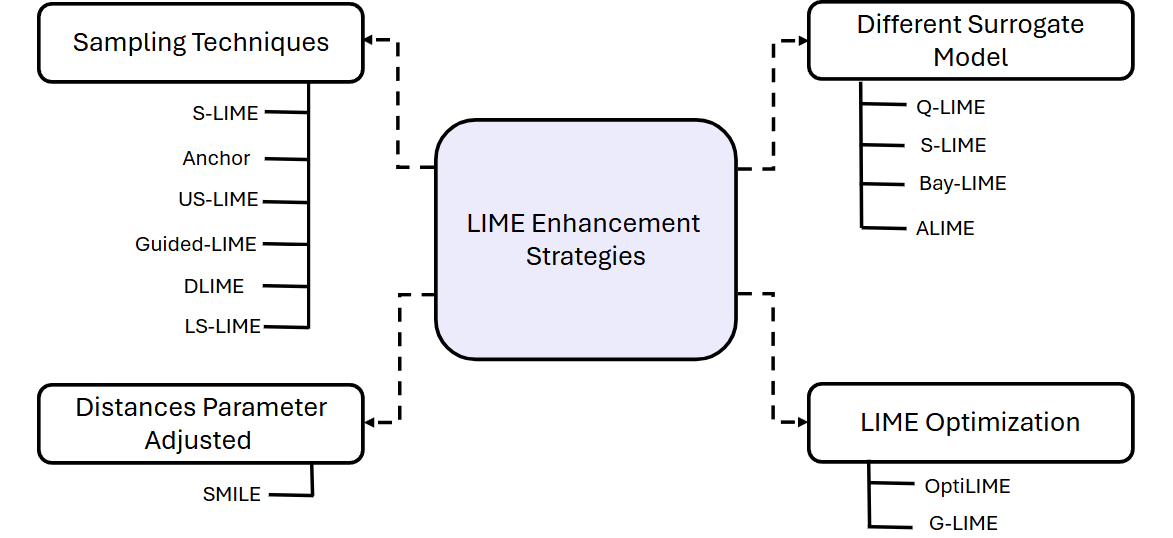}
    \caption{Taxonomy of LIME enhancement strategies: categorized into sampling techniques, surrogate model improvements, distance parameter adjustments, and optimization approaches.}
    \label{fig:LIME_methods}
\end{figure}

LIME has become one of the most widely used explainability techniques; however, over time, researchers have introduced various improvements to address its limitations and expand its applicability across different applications. As shown in Figure~\ref{fig:LIME_methods}, enhancements to LIME can be grouped into four main categories.

A key improvement in LIME's explanations is the refinement of distance measures used to weigh data points. For instance, SMILE uses statistical techniques to generate more consistent and reliable explanations. However, while these improvements enhance reliability, they also come with increased computational complexity. Another approach to improving LIME focuses on upgrading its simple linear models with more sophisticated alternatives. Methods such as Q-LIME~\cite{bramhall2020qlime}, S-LIME~\cite{zhou2021s}, Bay-LIME~\cite{zhao2021baylime}, and ALIME~\cite{shankaranarayana2019alime} use more advanced surrogate models to provide explanations that better reflect the underlying model's behaviour, capturing non-linear relationships that the original method might miss. Furthermore, researchers have explored optimisation strategies to enhance the efficiency and stability of LIME. Techniques like OptiLIME~\cite{visani2020optilime} and G-LIME~\cite{li2023g} work to strike a balance between accuracy and computational cost, ensuring that explanations remain reliable while keeping the process computationally manageable.

Recent research has extended LIME to support specific data types and improve its interpretability. Adaptations such as PointNet LIME for 3D point cloud data~\cite{levi2024fast}, TS-MULE for time series data~\cite{schlegel2021ts}, Graph LIME for graph-structured data~\cite{huang2022graphlime}, Sound LIME for audio data~\cite{mishra2017local}, and B-LIME for ECG signal data~\cite{abdullah2023b} demonstrate how LIME can be tailored to diverse applications. In addition, several modifications have been proposed to enhance LIME's approach, including the use of alternative surrogate models, fine-tuning distance parameters, optimising sampling techniques, and refining computational efficiency to improve both interpretability and accuracy.

Other extensions include DSEG-LIME~\cite{knab2024dseg}, which tailors LIME for segmentation models by generating perturbations at the object level rather than the pixel level, improving interpretability in image segmentation tasks. SLICE~\cite{bora2024slice} introduces structured sampling and clustering to ensure diverse and representative local explanations, reducing redundancy in LIME samples. Stratified LIME~\cite{rashid2024using} improves fidelity by stratifying the input space and generating more targeted local samples for different subpopulations. US-LIME~\cite{saadatfar2024us} employs uncertainty sampling to guide instance selection, increasing the reliability and faithfulness of local explanations. S-LIME~\cite{lam2025local} uses self-supervised learning to identify important regions in the feature space and provide richer explanations.

These methods aim to enhance interpretability across various domains, including segmentation, survival analysis, multimodal learning, and attention-based contexts. Additionally, a wide range of other extensions have been developed to enhance LIME’s capabilities across domains and explanation strategies, including: LSLIME, LIME-Aleph, Anchors~\cite{ribeiro2018anchors}. Numerous other LIME-based and post-hoc explanation variants have been proposed, encompassing a wide range of methodological innovations. These have been comprehensively reviewed and categorised in~\cite{knab2025lime}.

\subsection{Explainable AI (XAI) for Large Language Models}

Large Language Models (LLMs) have revolutionised natural language processing, enabling text generation, translation, and question answering tasks. However, their opaque ``black-box'' nature raises concerns regarding interpretability, fairness, and trustworthiness~\cite{zhao2024explainability, bender2021dangers}. Explainability in LLMs is crucial for ensuring accountability, enhancing user trust, and mitigating biases. This literature review explores existing research on explainability methods, applications, and challenges in LLMs.

The conceptual framework for explainability in LLMs includes four key dimensions: Post-hoc Explainability, Mechanistic Interpretability, Human-centric Explainability, and Applications and Challenges. These categories reflect the main approaches and practical considerations in enhancing the transparency and trustworthiness of LLMs.

After training, post-hoc explainability methods aim to interpret LLMs, providing insights into their decision-making processes~\cite{zhao2024explainability, madsen2022post}. Zhao et al.~\cite{zhao2024explainability} present a taxonomy of explainability techniques for LLMs, categorising approaches into local and global explanations, and discussing limitations and future research directions. Similarly, Bender et al.~\cite{bender2021dangers} highlight the risks of opaque LLMs, including biases and misinformation, and call for increased transparency and accountability. Recent surveys further expand this field by reviewing how LLMs contribute to explainable AI at large~\cite{zhao2024explainability}. Amara et al.~\cite{amara2025concept} recently introduced ConceptX, a concept-level attribution method highlighting semantically meaningful input phrases through a coalition-based Shapley framework, offering robust and human-aligned explanations.

Mechanistic interpretability seeks to uncover how LLMs process information internally~\cite{nanda2023progress, anthropic2024claude3}. 
Olah et al.~\cite{nanda2023progress} propose reverse-engineering techniques to dissect internal representations. 
Nanda et al.~\cite{nanda2023progress} examine how LLMs develop abstract representations and generalisation, introducing progress measures for interpretability research. Recent work by Anthropic~\cite{anthropic2024claude3} demonstrates the role of mechanistic interpretability in understanding and controlling model behaviour. 
Additionally, the SEER framework~\cite{chen2025seer} introduces self-explainability mechanisms to enhance the interpretability of internal representations in LLMs. 
The CELL framework~\cite{luss2024cell} further advances mechanistic interpretability by integrating concept-based explanations directly into the training and representation learning of language models.

Recent advancements from the Transformer Circuits framework have further enriched mechanistic interpretability. Using causal scrubbing, Ameisen et al.~\cite{ameisen2025circuit} introduced Attribution Graphs to uncover emergent structures within neural networks. In contrast, Nanda et al.~\cite{nanda2024monosemanticity} investigated the scaling of monosemanticity, extracting interpretable features from state-of-the-art models. Additional recent contributions explore specific tasks such as relevance estimation~\cite{elhage2021mathematical}, emotion inference~\cite{tak2025mechanistic}, and geospatial reasoning~\cite{de2025geospatial}. These works leverage techniques like activation patching and sparse autoencoders to reveal LLM internals. Sparse interpretability methods are further reviewed comprehensively in~\cite{elhage2021mathematical}. Neuroscientific perspectives have also been applied, using dynamical systems theory to model token-level trajectories within transformers~\cite{fernando2025transformer}.

Human-centric explainability focuses on making LLM outputs understandable to non-experts~\cite{ji2023survey, mondorf2024beyond, goethals2025if}. Ji et al.~\cite{ji2023survey} investigate hallucination in LLM outputs and propose techniques for detecting and mitigating incorrect responses. Krause et al.~\cite{mondorf2024beyond} assess reasoning abilities compared to human performance, and Martens et al.~\cite{goethals2025if} explore counterfactual explanations to improve human understanding. Recent works extend this perspective by enabling interactive explanations of black-box models~\cite{goethals2025if} and by benchmarking explanation quality in clinical domains~\cite{chen2025benchmarking}. These approaches improve the accessibility and transparency of LLM outputs in real-world decision-making.

Recent research has explored cost-efficient, model-agnostic explanation methods for LLMs~\cite{liu2025towards}, which align well with the objectives of gSMILE; their approach leverages sampling and scoring strategies to generate faithful explanations without requiring access to model internals.

Explainability enhances debugging, bias detection, regulatory compliance, and user trust in LLMs~\cite{adadi2018peeking}. Challenges remain, including defining ``meaning'' in LLM-generated outputs~\cite{adadi2018peeking}, managing the ethical and societal implications of black-box models, and balancing performance with interpretability. Recent work has demonstrated SMILE's applicability beyond language, such as in instruction-based image editing~\cite{dehghani2024mapping}, highlighting its modality-agnostic nature.

\section{Problem Definition}\label{sec:problem_definition}

This section introduces the core concepts and mathematical framework underlying our approach. We begin by defining the key notation and problem setup, followed by a formal description of how we quantify the impact of input perturbations. We then describe the use of a surrogate model to approximate model behaviour locally and provide theoretical justification based on the Wasserstein distance~\cite{ribeiro2016should, arjovsky2017wasserstein, peyre2019computational}. Lastly, we outline a structured methodology for implementing a Wasserstein-based surrogate framework tailored to large language models~\cite{molnar2020interpretable}.

Word Mover’s Distance (WMD) is the 1-Wasserstein (Earth Mover’s) distance between two documents represented as normalised bags-of-words in an embedding space, using the Euclidean distance between word embeddings as the ground cost~\cite{kusner2015word}. WMD provides a semantically meaningful measure of dissimilarity between documents by capturing the minimum amount of semantic ``movement'' needed to align their word distributions, making it well-suited for our interpretability framework.

\subsection{Notation and Problem Setting}

\paragraph{Setting the Stage.}
We aim to construct a local surrogate model that explains the behaviour of a black-box large language model (LLM), denoted as \( \pi^{(n)} \), in response to a natural language input prompt. The goal is to understand how specific parts of the input affect the model's output distribution. The following notation is used:

\begin{itemize}
    \item \textbf{Input:} \( x \in \mathcal{X} \), representing the original input prompt.
    \item \textbf{Perturbations:} \( \{\hat{x}_j\}_{j=1}^{J} \), a set of \( J \) perturbed versions of \( x \), generated via minor token-level edits.
    \item \textbf{Model Outputs:} \( \pi^{(n)}(y \mid \hat{x}_j) \), the model’s probability distribution over output space \( \mathcal{Y} \), conditioned on the perturbed input \( \hat{x}_j \). For the unperturbed input \( x \), we write \( \pi^{(n)}(y \mid x) \).
\end{itemize}

\subsection{Input-Level Distance}

To quantify the semantic difference between the original prompt \( x \) and each perturbed version \( \hat{x}_j \), we compute the Word Mover’s Distance (WMD)~\cite{kusner2015word}. We denote this input-level semantic distance as IWMD (Inner WMD):

\begin{equation}
\delta_{x_j} = \mathrm{IWMD}(x, \hat{x}_j),
\label{eq:input-distance}
\end{equation}

Where \( \delta_{x_j} \in \mathbb{R}_{\geq 0} \) denotes the semantic distance between the original and perturbed prompts.

\subsection{Output-Level Distribution Shift}

To assess the effect of input perturbation on the model’s behaviour, we compute the Wasserstein distance between the output distributions of the original and perturbed inputs. We refer to this as OWMD (Outcome WMD):

\begin{equation}
\mathrm{OWMD}(x, \hat{x}_j) = W \left( \pi^{(n)}(y \mid x), \pi^{(n)}(y \mid \hat{x}_j) \right),
\label{eq:owmd-definition}
\end{equation}

\begin{equation}
\Delta(x, \hat{x}_j) = \mathrm{OWMD}(x, \hat{x}_j),
\label{eq:output-shift}
\end{equation}

where \( y \in \mathcal{Y} \), and \( W(p, q) \) denotes the Wasserstein distance between two probability distributions \( p \) and \( q \).

\subsection{Weighting via Input Similarity}

Perturbations that are semantically closer to the original input are weighted more heavily. We use a Gaussian-based weighting scheme~\cite{ribeiro2016should}:

\begin{equation}
w_j = \exp \left( - \left( \frac{ \delta_{x_j} }{ \sigma } \right)^2 \right),
\label{eq:weights}
\end{equation}

Where \( w_j \in (0,1] \) denotes the relevance weight of \( \hat{x}_j \), and \( \sigma > 0 \) is a kernel width parameter that controls the rate of exponential decay.

\subsection{Fitting a Local Surrogate}

Each perturbed input \( \hat{x}_j \) is mapped to a feature vector \( z_j \in \mathbb{R}^d \). We fit a linear surrogate model of the form:

\begin{equation}
h_\theta(z_j) = \theta_0 + \theta^\top z_j,
\label{eq:surrogate}
\end{equation}

where \( \theta_0 \in \mathbb{R} \) is a bias term and \( \theta \in \mathbb{R}^d \) is the weight vector. Parameters \( \theta \) are estimated by minimising the weighted squared error.

\begin{equation}
\min_{\theta} \sum_{j=1}^{J} w_j \left( h_\theta(z_j) - \Delta(x, \hat{x}_j) \right)^2,
\label{eq:loss}
\end{equation}

Where the loss ensures local ATT fidelity between the surrogate and the observed output shifts.

\subsection{Theoretical Justification (Lipschitz Smoothness)}

\paragraph{Shift Function Definition.}

We define a shift function that locally quantifies the model's response to perturbation:

% \begin{equation}
% f(x_1, \hat{x}_1) = W \left( \pi^{(n)}(y \mid x_1), \pi^{(n)}(y \mid \hat{x}_1) \right).
% \label{eq:shift-function}
% \end{equation}

\begin{equation}
f(x_a, \hat{x}_a) = W \left( \pi^{(n)}(y \mid x_a), \pi^{(n)}(y \mid \hat{x}_a) \right),
\label{eq:shift-function}
\end{equation}

If \( \pi^{(n)} \) is Lipschitz continuous with respect to its input, then the function \( f \) is also Lipschitz~\cite{beliakov2007smoothing}:

% \begin{equation}
% \left| f(x_1, \hat{x}_1) - f(x_2, \hat{x}_2) \right| \leq L \cdot \left( \|x_1 - x_2\| + \|\hat{x}_1 - \hat{x}_2\| \right),
% \label{eq:lipschitz}
% \end{equation}
\begin{equation}
\left| f(x_a, \hat{x}_a) - f(x_b, \hat{x}_b) \right| \leq L \times \left( \|x_a - x_b\| + \|\hat{x}_a - \hat{x}_b\| \right), \forall a,b \in \{1, \dots, N_{pert}\}
\label{eq:lipschitz}
\end{equation}

%The notation \( x_1 \) and \( \hat{x}_1 \) represents a specific input and its corresponding perturbation. The subscript allows us to extend the formulation to multiple input pairs \( (x_i, \hat{x}_i) \) when analysing the Lipschitz continuity or stability across perturbations.

The notation \( x_a \) and \( \hat{x}_a \) represents a specific input and its corresponding perturbation. The subscript allows us to extend the formulation to multiple input pairs \( (x_i, \hat{x}_i) \) when analysing the Lipschitz continuity or stability across perturbations. \( N_{pert} \) denotes the total number of perturbations. 

%Note that the term \( L \cdot (...) \) simply represents the Lipschitz constant multiplied by the sum of input differences, as measured by the selected norm.

where \( L > 0 \) is the Lipschitz constant, and \( \|\cdot\| \) denotes a norm over the input space (e.g., token embedding space).

\paragraph{Relation to Total Variation.}

For two distributions \( p \) and \( q \) over output space \( \mathcal{Y} \), the Kantorovich–Rubinstein duality yields:

\begin{equation}
W(p, q) \leq \text{diam}(\mathcal{Y}) \cdot \mathrm{TV}(p, q),
\label{eq:wasserstein-tv}
\end{equation}

Where $\text{diam}(\mathcal{Y}) = \sup_{y_1, y_2 \in \mathcal{Y}} d(y_1, y_2)$ denotes the \emph{diameter} of the output space under the chosen ground metric $d$, i.e., the maximum possible distance between any two outputs.

The total variation distance ($TV$) is given by:

\begin{equation}
\mathrm{TV}(p, q) = \frac{1}{2} \sum_{y \in \mathcal{Y}} |p(y) - q(y)|.
\label{eq:tv}
\end{equation}

This bound highlights that small changes in distribution (in the TV sense) imply a bounded Wasserstein distance.

\subsection{Linear Surrogates Approximate Lipschitz Functions}

Due to the differentiability of Lipschitz functions, the shift function \( f \) can be locally approximated by a first-order Taylor expansion:

\begin{equation}
f(x') \approx f(x) + \nabla f(x)^\top (x' - x),
\label{eq:linear-approximation}
\end{equation}

where \( \nabla f(x)^\top (x' - x) \) denotes the dot product between the gradient of \( f \) and the local displacement in input space.

The complete Wasserstein-based LIME surrogate procedure consists of generating perturbations $\{\hat{x}_j\}_{j=1}^{J}$ near input prompt $x$. The process then computes semantic distances $\delta_{x_j}$ (Equation~\ref{eq:input-distance}) and output shifts $\Delta(x, \hat{x}_j)$ (Equation~\ref{eq:output-shift}). Following this, similarity weights $w_j$ are assigned via Equation~\ref{eq:weights}. The procedure continues by extracting feature representations $z_j$ for each perturbation and fitting a local surrogate model $h_\theta$ by minimising the loss (Equation~\ref{eq:loss}). Finally, the fitted model $h_\theta$ is used to interpret the influence of each input feature on model behaviour. This framework offers a principled and interpretable approach to analysing large language models through localised perturbation and response analysis.

\section{Proposed Method}

The proposed approach, named \textbf{gSMILE} (Generative SMILE), enhances the interpretability of large language models (LLMs) for text generation by analysing how specific textual inputs influence the generated outputs. Understanding the interaction between input prompts and generated text improves transparency and predictability in the model's behavior~\cite{bommasani2021opportunities, molnar2020interpretable}.

\begin{figure}[H]
    \centering
    \includegraphics[width=0.9\linewidth]{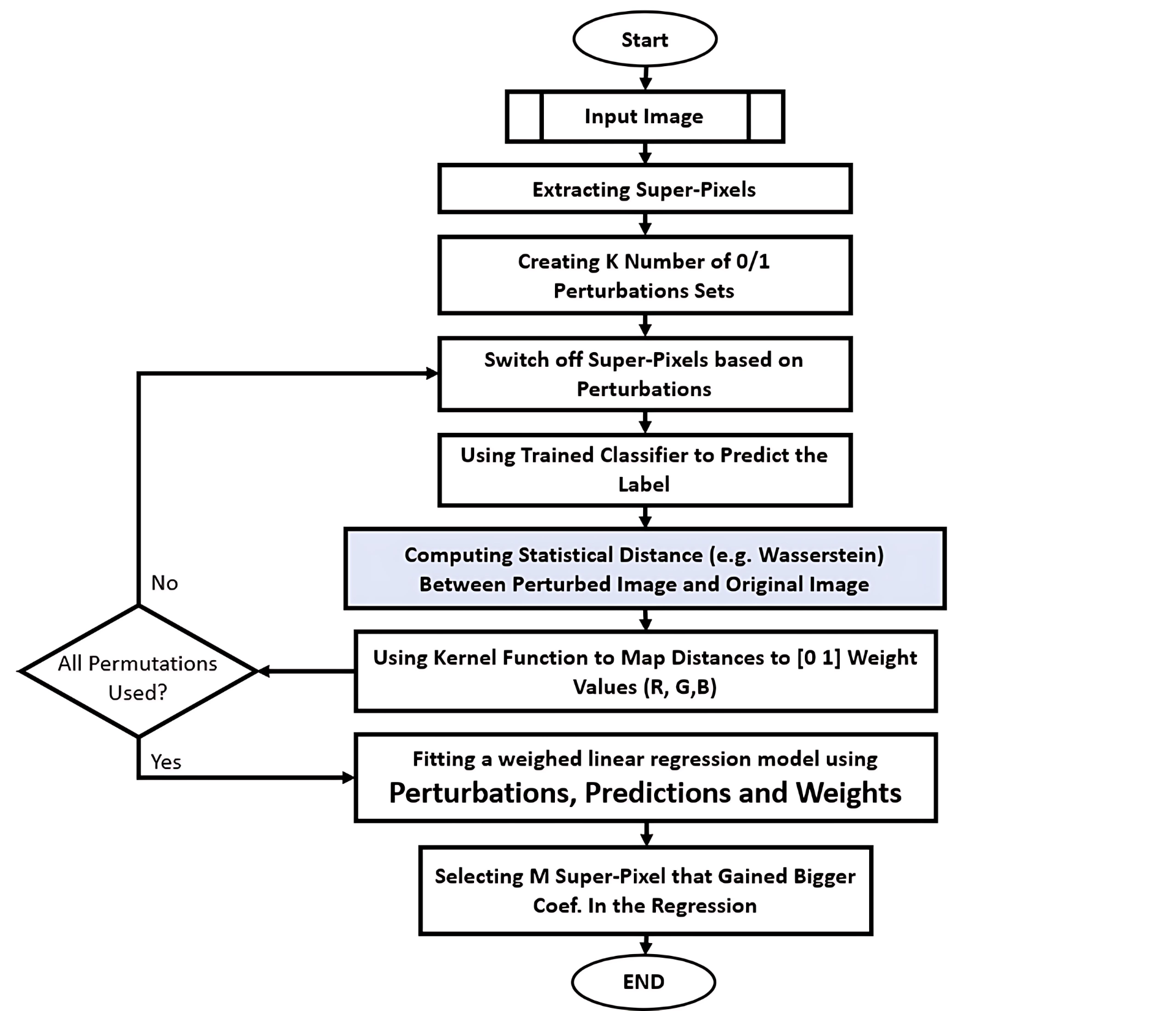}
    \caption{SMILE flowchart for explaining image classification~\cite{aslansefat2023explaining}}
    \label{fig:smile}
\end{figure}

In the context of classification tasks, Fig.~\ref{fig:smile} illustrates SMILE, a tool designed to explain model predictions by isolating critical features that drive decision-making~\cite{aslansefat2023explaining}. SMILE segments an input (e.g., an image) into meaningful components: ``super-pixels'' for images or ``logical sections'' for text, and creates perturbations by selectively modifying these components. These perturbations reveal how different input elements contribute to the final prediction.

SMILE records the model's predictions for each perturbed input. It calculates distance scores between the original and modified outputs using metrics such as the Wasserstein distance~\cite{arjovsky2017wasserstein, peyre2019computational}. These distances are transformed into similarity-based weights using a kernel function, enabling a balanced weighting scheme. The results are used to train a weighted linear regression model that identifies the most influential input components.

\begin{figure}[H]
    \centering
    \includegraphics[width=0.8\linewidth]{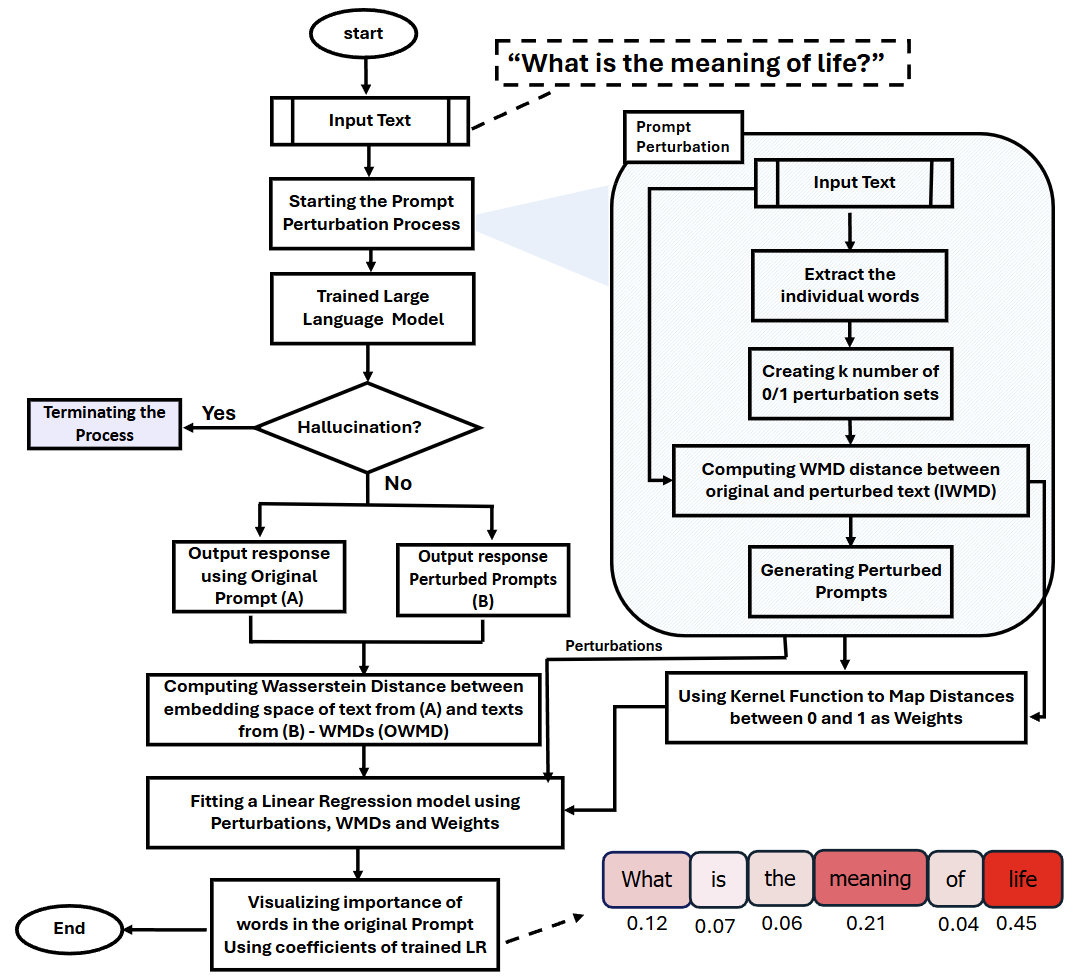}
    \caption{gSMILE flowchart for Large Language Models}
    \label{fig:smile_llm}
\end{figure}

Inspired by SMILE, we propose a similar method for interpreting text prompt-based text generation models, as illustrated in Fig.~\ref{fig:smile_llm}. Instead of perturbing the generated image, as in the original SMILE framework, we perturb the input prompt. Each perturbed prompt $\hat{x}_j$ produces a corresponding text output, from which we compute the outcome-level semantic shift $\Delta(x, \hat{x}_j)$ using the IWMD as defined ~\ref{eq:input-distance}. This allows us to identify which words in the input prompt most influence the model’s output.

The set of output shifts $\{\Delta(x, \hat{x}_j)\}_{j=1}^J$ is then used as the outcome variable in a weighted linear regression model, with weights $w_j$ computed from the input-level semantic distances $\delta_{x_j}$ via a Gaussian kernel (Eq.~\ref{eq:weights}). This regression estimates how each input feature (token) influences the generated output within the local neighbourhood of $x$. The resulting coefficients $\theta$ quantify the relative importance of each token, which we visualise as a heatmap highlighting the most influential words in the prompt.

Using this enhanced three-step process, which builds on SMILE's methodology~\cite{aslansefat2023explaining}, we aim to help users understand how particular elements of text prompts impact text generation. This approach improves user control and predictability by showing which textual components, such as stylistic instructions or descriptive terms, most influence the final output.

In the first step, we employ a black-box text generation model $\pi^{(n)}$ that receives an original prompt $x$ and produces an associated output $\pi^{(n)}(y \mid x)$. We then construct a set of perturbed prompts $\{\hat{x}_j\}_{j=1}^J$ by applying small, semantically meaningful edits to $x$ (e.g., removing or replacing specific tokens). For each $\hat{x}_j$, the model generates $\pi^{(n)}(y \mid \hat{x}_j)$. Ideally, these outputs remain similar to the original output, with variation arising primarily from the altered components of the prompt~\cite{zhang2024comprehensive}.

We quantify the distributional shift between the original and perturbed outputs using the OWMD as defined~\ref{eq:owmd-definition} $\Delta(x, \hat{x}_j)$, as defined in Eq.~\ref{eq:output-shift}, computed in the output embedding space~\cite{rubner2000Word, arjovsky2017wasserstein}. This metric is chosen for its ability to capture both broad and subtle differences compared to ECDF-based measures~\cite{peyre2019computational}.

Finally, we fit a weighted linear surrogate model $h_\theta$ (Eq.~\ref{eq:surrogate}) by minimising the loss in Eq.~\ref{eq:loss}, linking the feature representation $z_j$ of each perturbed prompt to its corresponding $\Delta(x, \hat{x}_j)$. The learned coefficients $\theta$ provide an interpretable mapping between input prompt components and changes in the model's generated output~\cite{murdoch2019interpretable, hastie2009elements}.

\subsection{Text Generating and Perturbed Prompts}

We begin by generating variations of the original text prompt $x$, referred to as ``perturbed prompts'' $\{\hat{x}_j\}_{j=1}^J$, to analyse how subtle changes in the input affect the output of a text generation model~\cite{li2023guiding}. The original prompt $x$ is first tokenised into individual tokens. From these tokens, multiple perturbed prompts are constructed by selectively including, excluding, or substituting specific words or phrases. Each perturbed prompt $\hat{x}_j$ is then passed to the text generation model $\pi^{(n)}$, which produces a corresponding output. The output generated from the original prompt, $\pi^{(n)}(y \mid x)$, serves as the reference (or baseline) for comparison.

Formally, the outputs for the original and perturbed prompts follow the notation in Section~\ref{sec:problem_definition}:

the model generates $y_{\text{org}} \sim \pi^{(n)}(y \mid x)$ for the original prompt, and $y_{\text{pert}, j} \sim \pi^{(n)}(y \mid \hat{x}_j)$ for each perturbation $j \in \{1, \dots, J\}$. 

Here, $\pi^{(n)}$ denotes the black-box text generation model, $y_{\text{org}}$ is the output for the original prompt, and $y_{\text{pert}, j}$ denotes the output for the $j$-th perturbation. These outputs are subsequently used in the computation of the output shift $\Delta(x, \hat{x}_j)$ (Eq.~\ref{eq:output-shift}) to measure how individual token-level changes in the prompt influence the model’s behavior~\cite{ribeiro2016should, zhu2023promptbench}.

\subsection{Creating the Interpretable Space}

To represent the output space of the text generation model in a one-dimensional similarity space, we compute the OWMD distance between the output for each perturbed prompt $\hat{x}_j$ and the output for the original prompt $x$ (as defined in Eq.~\ref{eq:output-shift})~\cite{arjovsky2017wasserstein, peyre2019computational}. Throughout this process, the input structure remains unchanged, and only the perturbed tokens differ across inputs.

To capture semantic similarity between generated outputs, we employ the Outcome Word Mover’s Distance (OWMD) to measure distances between texts based on their representations in a semantic embedding space~\cite{kusner2015word}. Operating in this embedding space focuses the comparison on high-level semantic content, thereby mitigating lexical noise and improving the reliability of the computed distances. The resulting scalar values $\Delta(x, \hat{x}_j)$ are then used as the dependent variable in the weighted regression model.

Here, $\phi(\cdot)$ denotes the output embedding function, $n$ is the dimensionality of the embedding space, and $P$ specifies the norm order used in the Wasserstein metric.

For each perturbation $j$, $\Delta(x, \hat{x}_j)$ measures the distributional shift in the generated output relative to the baseline.

After computing $\Delta(x, \hat{x}_j)$ for each perturbation, we assess their statistical significance using p-values~\cite{wasserman2004all}. This ensures that only meaningful output shifts are considered in the regression model. A p-value quantifies the probability that an observed Outcome Word Mover’s Distance (OWMD) between outputs could have occurred purely by random chance, under the null hypothesis that the perturbation has no real effect.

To control the sensitivity of \texttt{OWMD} to different types of changes, we use the \textbf{Wasserstein-$P$ metric}~\cite{peyre2019computational}. The norm order $P$ determines the scale of differences captured: smaller $P$ values emphasise fine-grained, local variations, whereas larger $P$ values capture broader, global shifts. By experimenting with different $P$ values, we select the one that best captures the semantic effects of prompt perturbations, thereby improving both the robustness and interpretability of the analysis.

We compute p-values for each observed \texttt{OWMD} using a Bootstrap resampling procedure~\cite{efron1994introduction, dehghani2024mapping}, as detailed in Algorithm~\ref{alg:bootstrap}. This non-parametric approach repeatedly resamples the combined set of embeddings from both the original and perturbed outputs to construct an empirical distribution of \texttt{OWMD} values under the null hypothesis. The p-value is then estimated as the proportion of bootstrap distances greater than the observed \texttt{OWMD}. Perturbations with high p-values, indicating non-significant differences, are excluded from further interpretation.

\begin{algorithm}[H]
\caption{Bootstrap-based P-Value Calculation for Outcome Wasserstein Distance (OWMD)}
\label{alg:bootstrap}
\SetAlgoLined
\KwIn{Two sets of text embeddings: $X$ and $Y$}
\KwOut{Observed Outcome Wasserstein Distance ($\texttt{OWMD}$) and p-value ($\texttt{pVal}$)}

$\texttt{MaxItr} \leftarrow 10^5$\;
$\texttt{OWMD} \leftarrow \texttt{Wasserstein\_Dist}(X, Y)$\;
$\texttt{XY} \leftarrow X \cup Y$\;
$L_X \leftarrow |X|$, $L_Y \leftarrow |Y|$\;
$\texttt{bigger} \leftarrow 0$\;

\For{$i \leftarrow 1$ \KwTo $\texttt{MaxItr}$}{
    Sample $e$ of size $L_X$ from $\texttt{XY}$ without replacement\;
    Sample $f$ of size $L_Y$ from $\texttt{XY}$ without replacement\;
    $\texttt{bootOWMD} \leftarrow \texttt{Wasserstein\_Dist}(e, f)$\;
    \If{$\texttt{bootOWMD} > \texttt{OWMD}$}{
        $\texttt{bigger} \leftarrow \texttt{bigger} + 1$\;
    }
}
$\texttt{pVal} \leftarrow \frac{\texttt{bigger}}{\texttt{MaxItr}}$\;
\Return{$\texttt{OWMD}, \texttt{pVal}$}\;
\end{algorithm}

When generating perturbed prompts from the original prompt $x$, preserving key terms is essential for maintaining semantic similarity between each perturbed prompt $\hat{x}_j$ and the source. To capture this similarity, we compute the \emph{input-level semantic distance} $\delta_{x_j}$ using the Inner Wasserstein Distance over tokens (IWMD)~\cite{kusner2015word}, as introduced in Eq.~\ref{eq:input-distance}. This value is later transformed into a sample weight $w_j$ for the interpretable surrogate model (Eq.~\ref{eq:weights}).

To incorporate input similarity into the regression model, each IWMD value is converted into a similarity weight via a Gaussian kernel, following Eq.~\ref{eq:weights}:

\begin{equation}
w_j(x, \hat{x}_j) = \exp \left[ - \left( \frac{\mathrm{IWMD}(x, \hat{x}_j)}{\sigma} \right)^2 \right],
\quad \forall j \in \{1, \dots, J\}.
\end{equation}

Here, $\sigma > 0$ is the kernel width parameter that controls the rate of decay in weight as semantic dissimilarity increases. The resulting weights $w_j \in (0, 1]$ assign greater importance to perturbed prompts that are more semantically similar to the original prompt $x$, and down-weight those that differ substantially.

\subsection{Developing the Interpretable Surrogate Model}

To interpret the influence of individual tokens on the model's output, we map the output space of the black-box text generation model $\pi^{(n)}$ to a one-dimensional distance scale, where each point corresponds to a perturbed prompt~\cite{habib2024exploring}. This mapping enables the training of a local surrogate model that relates structural changes in the input prompt to variations in output behaviour.

In this weighted linear regression framework, the independent variables are binary (or one-hot) perturbation vectors indicating which tokens from the original prompt $x$ are retained or altered in $\hat{x}_j$. The dependent variable is the output-level distribution shift $\Delta(x, \hat{x}_j)$, defined in Eq.~\ref{eq:output-shift} as the Wasserstein distance between the output distributions for $x$ and $\hat{x}_j$. Each perturbation is assigned a similarity-based weight $w_j$ from Eq.~\ref{eq:weights}, computed via the Gaussian kernel over the input-level distance $\delta_{x_j}$.

The surrogate model parameters are estimated by minimising the weighted squared error:
\begin{equation}
\label{eq:eq5}
\mathcal{L}(\pi^{(n)}, \delta_{x_j}, x) =
\frac{1}{J} \sum_{j=1}^{J} w_j(x, \hat{x}_j) 
\cdot \left( \Delta(x, \hat{x}_j) - h_\theta(z_j) \right)^2
\end{equation}

The fitted coefficients $\theta$ quantify the contribution of each token to the observed output shift, enabling interpretable attribution.

We evaluate the proposed attribution method using several state-of-the-art large language models: OpenAI’s \texttt{gpt-3.5-turbo-instruct}~\cite{ye2023comprehensive}, Anthropic’s Claude 2.1~\cite{claude2024}, and Meta’s LLaMA 3.1 Instruct Turbo (70B)~\cite{llama2024huggingface}.

\subsection{Evaluation Metrics}

\begin{figure}[H]
    \centering
    \includegraphics[width=1\linewidth]{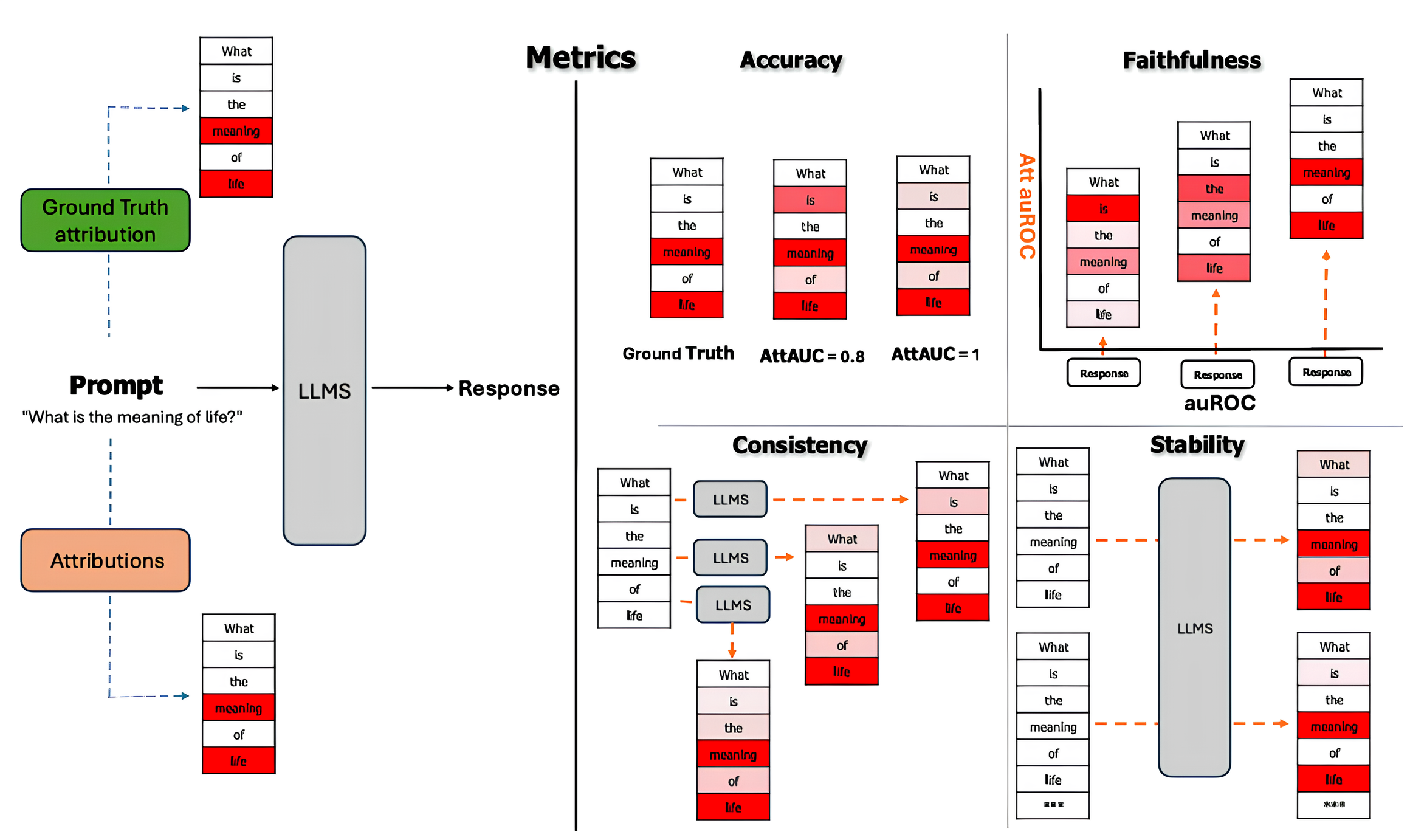}
    \caption{Given a prompt (e.g., “What is the meaning of life?”), We compare ground truth attributions with model-generated attributions for input tokens. Attribution quality is evaluated using four metrics: ATT Accuracy quantifies alignment of attributions with ground-truth (via AttAUC scores). ATT faithfulness measures whether attributions reflect the importance of tokens for the model’s response (via auROC). ATT consistency assesses whether results remain stable across different model runs or hyperparameters. ATT stability examines the sensitivity of attributions to small perturbations in the input prompt.}
    \label{fig:LIMEDEV}
\end{figure}

In our investigation, we adopt a suite of evaluation metrics inspired by foundational work provided by Google~\cite{sanchez2020evaluating}, emphasising the multifaceted nature of assessing explainable models. This work highlights the significance of metrics such as ATT accuracy, ATT stability, ATT fidelity, ATT faithfulness, and ATT consistency as essential tools for a rigorous evaluation of model behaviour, particularly when comparing explainable models to traditional black-box models, as you can see in Fig.~\ref{fig:LIMEDEV}. The adoption of these metrics provides a structured methodology to dissect and understand model reliability in a more holistic manner~\cite{sanchez2020evaluating}.

\subsubsection{Attribution Accuracy}

\textbf{Att Accuracy} measures how well the model's output aligns with the expected results (ground truth). Specifically, we compare the model's attribution scores (attention to text elements) against ground truth labels that identify the most relevant text elements for specific generated text segments~\cite{fong2017interpretable}.

To quantify this, we use the \textbf{Area Under the Curve (AUC)} of the Receiver Operating Characteristic (ROC)~\cite{hanley1982meaning}. AUC reflects the model's ability to rank relevant elements (from the ground truth) higher than irrelevant ones:
\begin{itemize}
    \item \textbf{AUC $\sim$1:} The model effectively distinguishes relevant elements, closely matching human-identified ground truth.
    \item \textbf{AUC $\sim$0.5:} The model's ranking is random, showing poor alignment with the ground truth.
\end{itemize}

For example, Fig.~\ref{fig:LIMEDEV} shows that the ground truth identifies the words \emph{“meaning”} and \emph{“life”} as the most relevant text elements for generating the corresponding text output. The metric \textbf{AttAUC} evaluates the model's alignment with the ground truth, where:

\begin{itemize}
    \item \textbf{AttAUC = 1.0:} Perfect alignment.
    \item \textbf{Lower AttAUC (0.8):} Less accurate~\cite{powers2020evaluation}.
\end{itemize}

The heatmaps produced by the model visually represent these attention scores, as shown in Fig.~\ref{fig:LIMEDEV}, where darker red shades indicate higher attribution to specific text elements. An accurate model should correctly assign higher attention to the text elements that most influence the generated text output, improving the interpretability and reliability of the model~\cite{fong2017interpretable}.

\subsubsection{Attribution Faithfulness}
ATT faithfulness refers to the extent to which an explanation accurately reflects a model’s reasoning process. A faithful attribution assigns high importance to input tokens that genuinely drive the model’s output, rather than highlighting spurious correlations or irrelevant features. This property is widely recognised as essential for trustworthy interpretability in natural language processing and other domains \cite{lyu2024towards, agarwal2024faithfulness}.  

In our study, we evaluate ATT faithfulness through correlation analysis between attribution-based metrics (ATT ACC, ATT F1, ATT AUROC) and externally reported benchmark accuracies of the models being assessed, including OpenAI’s \texttt{gpt-3.5-turbo-instruct}, Meta’s LLaMA~3.1 Instruct Turbo (70B), and Anthropic’s Claude~2.1 \cite{gpt35_mmlu_leaderboard, valsai2025llama70, anthropic2023claude2modelcard}.

Formally, we compute the Pearson correlation coefficient:
\begin{equation}
r = \frac{\sum_{i=1}^{n} \bigl(x_i - \bar{x}\bigr)\bigl(y_i - \bar{y}\bigr)}{\sqrt{\sum_{i=1}^{n} (x_i - \bar{x})^2}\,\sqrt{\sum_{i=1}^{n} (y_i - \bar{y})^2}},
\label{eq:pearson}
\end{equation}
where $x_i$ are published benchmark ATT Accuracy values, $y_i$ are the corresponding attribution scores, and $\bar{x}, \bar{y}$ denote their means. A strong positive $r$ indicates that attribution metrics faithfully track external measures of model competence.  

Thus, ATT faithfulness in our framework is treated as an externally validated property: explanations are considered reliable to the extent that attribution scores correlate with established benchmarks, ensuring interpretability methods provide meaningful insights into large language models, particularly in high-stakes applications \cite{jacovi2020towards}.

\subsubsection{Attribution Stability}

ATT stability in an explainable model refers to its ability to provide consistent explanations even when minor changes are made to the input data~\cite{samek2017explainable}. In other words, small perturbations in the input should not significantly affect the model's predictions or explanations. ATT stability is a key property for fostering trust in explainable artificial intelligence or XAI systems~\cite{doshi2017towards}. It reassures users that the model's interpretations are not arbitrary or overly sensitive to irrelevant variations~\cite{lipton2018mythos}.

In the example text, the attention scores are visualised for two slightly different text prompts. The heatmaps show how the model assigns attention to each word. A stable model, as demonstrated in Fig.~\ref{fig:LIMEDEV}, assigns similar importance to the words \textit{make} and \textit{rainy} in both prompts, even though the token \texttt{\*\*\*} was added in one case. This ATT consistency is critical for ensuring the model behaves predictably across minor input variations~\cite{ahmadi2024explainability}. Without such ATT stability, the explanations generated by the model could lead to confusion or misinterpretation, especially in high-stakes scenarios such as medical diagnosis or financial decision-making~\cite{ribeiro2016should}.

We use the \textbf{Jaccard index} to quantify ATT stability, which measures the similarity between two sets of model predictions. For two sets \( A \) and \( B \), the Jaccard index is calculated as follows in Eq.~\ref{eq:jaccard}:

\begin{equation}
\label{eq:jaccard}
\text{Jaccard}(A, B) = \frac{|A \cap B|}{|A \cup B|}
\end{equation}

This metric reflects the proportion of shared elements between the sets, where values closer to 1 indicate greater ATT stability and similarity in explanations~\cite{samek2017explainable}. A higher Jaccard index confirms the model's ability to yield stable outputs, thus providing consistent and reliable interpretations~\cite{vaswani2017attention}. By ensuring ATT stability, we not only enhance the robustness of the model but also its usability and credibility in real-world applications~\cite{doshi2017towards}.

\subsubsection{Attribution consistency}

In evaluating the ATT consistency of our model, we assess whether it produces stable and reliable outputs when the same input is provided multiple times. ATT consistency ensures that the model behaves predictably, reinforcing trust in its outputs and usability across diverse scenarios~\cite{ribeiro2016should}.

As shown in Fig.~\ref{fig:LIMEDEV}, when the input phrase ``what is the meaning of life?'' was fed into the model multiple times, the output remained stable across iterations. This ATT consistency is crucial for applications where repeatable and reliable results are required. In our case, the model consistently assigns higher weights to the words ``meaning'' and ``life'', demonstrating that it understands the importance of these words in generating the desired output~\cite{vaswani2017attention}. Such behaviour underscores the model's ability to focus on the semantic core of the input, a trait essential for maintaining coherence in practical use cases~\cite{lipton2018mythos}.

In practical applications, such as automated content generation, interactive systems, or scenario simulations, users depend on models to produce results that do not fluctuate unexpectedly. Any inconsistencies could undermine user trust and compromise the system's overall effectiveness. Our model enhances user confidence by demonstrating consistent outputs and establishes a foundation for further optimisation and integration into more complex systems. This ability to maintain ATT consistency across iterations directly contributes to the robustness and scalability of the model, making it well-suited for deployment in dynamic real-world environments~\cite{doshi2017towards}.

\subsubsection{Attribution Fidelity}

ATT fidelity is a metric used to evaluate how closely an explainable model replicates the behaviour of a black-box model. Specifically, the predictions of the black-box model, denoted as \( f(X_j) \), are compared to those of the explainable model, \( g(X_j) \), for each input perturbation \( j \). A variety of accuracy-based and error-based metrics are employed to assess this alignment, thereby quantifying how well the explainable model approximates the responses of the black-box model.

Let \( N_p \) denote the total number of perturbation samples (i.e., data points). One of the most commonly used accuracy metrics is the \textbf{coefficient of determination} \( R^2 \), which measures how well the explainable model’s predictions correlate with those of the black-box model~\cite{draper1998applied}. It is defined as:

\begin{equation} \label{eq:coefficient_of_determination}
R^2 = 1 - \frac{\sum_{j=1}^{N_p} (f(X_j) - g(X_j))^2}{\sum_{j=1}^{N_p} (f(X_j) - \bar{f})^2}
\end{equation}

where \( \bar{f} \) represents the mean of the black-box model's predictions \( f(X_j) \). A value of \( R^2 \) close to 1 indicates a strong alignment between the models.

In cases where data samples have different levels of importance, a weighted variant is used. The \textbf{weighted coefficient of determination} \( R_w^2 \) is given by~\cite{ahmadi2024explainability, biometrics19763211hocking1976analysis}:

\begin{equation} \label{eq:weighted_coefficient_of_determination}
R_w^2 = 1 - \frac{\sum_{j=1}^{N_p} (f(X_j) - g(X_j))^2}{\sum_{j=1}^{N_p} (f(X_j) - \bar{f}_w)^2}
\end{equation}

Here, \( \bar{f}_w \) is the weighted mean of the black-box predictions. These accuracy-based metrics (\( R^2 \) and \( R_w^2 \)) provide insight into how closely the explainable model aligns with the black-box model.

To account for both sample size and model complexity (i.e., number of features used), the \textbf{weighted adjusted coefficient of determination} \( \hat{R}_w^2 \) is used~\cite{ahmadi2024explainability}:

\begin{equation} \label{eq:weighted_adjusted_coefficient_of_determination}
\hat{R}_w^2 = 1 - (1 - R_w^2) \left[ \frac{N_p - 1}{N_p - N_s - 1} \right]
\end{equation}

Where \( N_s \) is the number of features in the interpretable model.

In addition to correlation-based metrics, error-based metrics are widely used to quantify differences in prediction values. The classical unweighted forms \textbf{Mean Squared Error (MSE)} and \textbf{Mean Absolute Error (MAE)} are:

\begin{equation} \label{eq:mse}
\text{WMSE} = \frac{\sum_{j=1}^{N_p} w_j \cdot (f(X_j) - g(X_j))^2}{\sum_{j=1}^{N_p} w_j}
\end{equation}

\begin{equation} \label{eq:mae}
\text{WMAE} = \frac{\sum_{j=1}^{N_p} w_j \cdot |f(X_j) - g(X_j)|}{\sum_{j=1}^{N_p} w_j}
\end{equation}

Here, \( w_j \) denotes the sample weight associated with the \( j \)-th perturbation.

Moreover, the \textbf{mean \( L_1 \)} and \textbf{mean \( L_2 \)} norms are also applied to evaluate absolute and squared differences respectively~\cite{goodfellow2016deep, ahmadi2024explainability}:

\begin{equation} \label{eq:l1_loss}
L_1 = \frac{1}{N_p} \sum_{j=1}^{N_p} |f(X_j) - g(X_j)|
\end{equation}

\begin{equation} \label{eq:l2_loss}
L_2 = \frac{1}{N_p} \sum_{j=1}^{N_p} (f(X_j) - g(X_j))^2
\end{equation}

These metrics, MSE, MAE, WMSE, WMAE, \( L_1 \), and \( L_2 \), together quantify how accurately the explainable model reproduces the black-box model’s predictions and reflect the risk of prediction discrepancies across the perturbation space.

We compute ATT fidelity across various scenarios by comparing the explainable model's predictions to those of the black-box text generation model. This includes analysing how different input perturbations affect ATT fidelity scores and comparing the performance of various distance metrics and surrogate models, such as linear regression and Bayesian ridge regression, in aligning with the black-box behaviour.

\begin{figure}[H]
    \centering
    \includegraphics[width=1\linewidth]{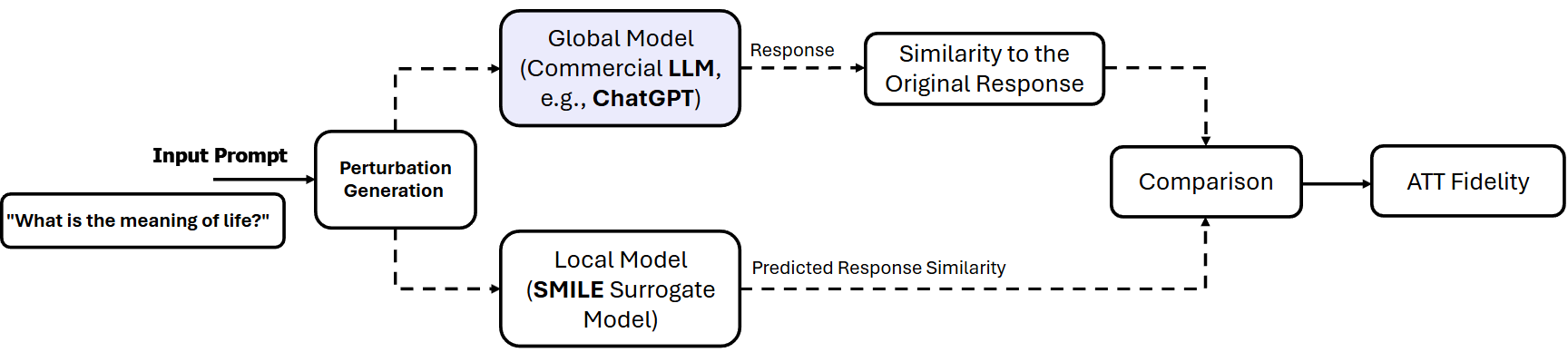}
    \caption{ATT fidelity evaluation framework. The input prompt is perturbed, 
    and both a Global Model (e.g., ChatGPT) and a Local Surrogate Model (SMILE) 
    generate similarity signals. Their comparison quantifies how closely 
    the surrogate approximates the black-box model’s behaviour.}
    \label{fig:fidelity}
\end{figure}

The overall workflow for computing ATT fidelity is shown in Figure~\ref{fig:fidelity}. 
As depicted, input prompts are perturbed, and both the Global Model and the Local Surrogate Model 
produce similarity signals that are subsequently compared to evaluate ATT fidelity.

\section{Experimental Results}

This section evaluates the proposed method's ability to improve explainability in instruction-based text generation, focusing on both performance and its implications for model transparency and usability. We analyse how prompt structure influences interpretability, how phrasing variations affect the clarity of explanations, and how the method performs across diverse scenarios.

For researchers, the analysis highlights linguistic factors that most affect interpretability; for practitioners, it offers actionable insights for prompt engineering and model debugging. Experiments were conducted using a range of datasets and three state-of-the-art text generation models, OpenAI’s \texttt{gpt-3.5-turbo-instruct}, Meta’s LLaMA 3.1 Instruct Turbo (70B), and Anthropic’s Claude 2.1, ensuring a robust and generalisable evaluation across architectures.

\subsection{Qualitative Results}

We assess interpretability through controlled variations of input prompts, including changes in sentence complexity, rephrasing, and the use of domain-specific terminology. This approach allows us to examine whether attribution patterns remain stable and meaningful under different linguistic conditions, a key requirement for trustworthy explainability tools. Observing these changes provides practical guidance for designing effective prompts and identifying potential reliability issues before deployment.

\subsubsection{Input Prompt Explainability}

The framework applies multiple prompt variations, such as rewording questions, adding descriptive modifiers, or altering word choice. It generates heatmaps showing the contribution of each token to the model’s output, with colour intensity indicating importance scores.

\begin{figure}[H]
    \centering
    \includegraphics[width=0.82\textwidth]{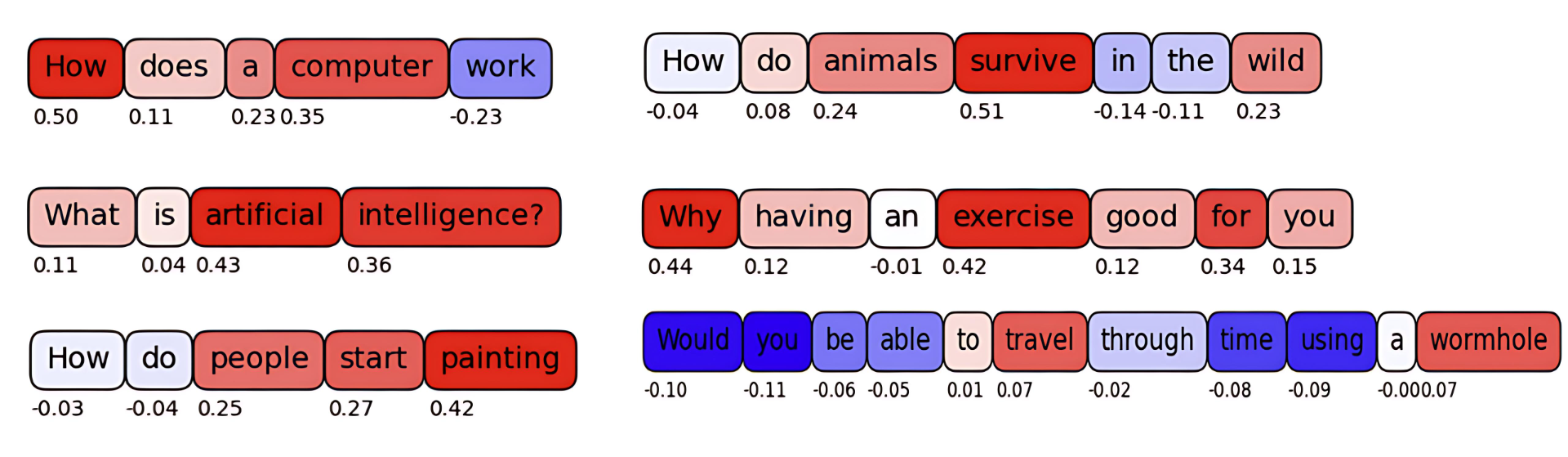}
    \caption{Text heatmaps showing word-level importance for various input prompts. The colour intensity represents the influence of each word on the model's output.}
    \label{fig:prompt_heatmaps}
\end{figure}

Figure~\ref{fig:prompt_heatmaps} illustrates how small changes in prompt phrasing or focus can lead to noticeably different attribution patterns when using OpenAI’s \texttt{gpt-3.5-turbo-instruct}. These differences are not merely cosmetic: they reveal how the model’s attention can shift even for semantically similar inputs. Understanding such sensitivity helps practitioners design prompts that minimise unwanted variation, while also highlighting cases where model behaviour may be unstable, valuable both for prompt-engineering strategies and for reliability assessments in real-world applications.

\subsubsection{Gender Fairness via Token-Level Attributions}

\begin{figure}[H]
    \centering
    \includegraphics[width=0.85\linewidth]{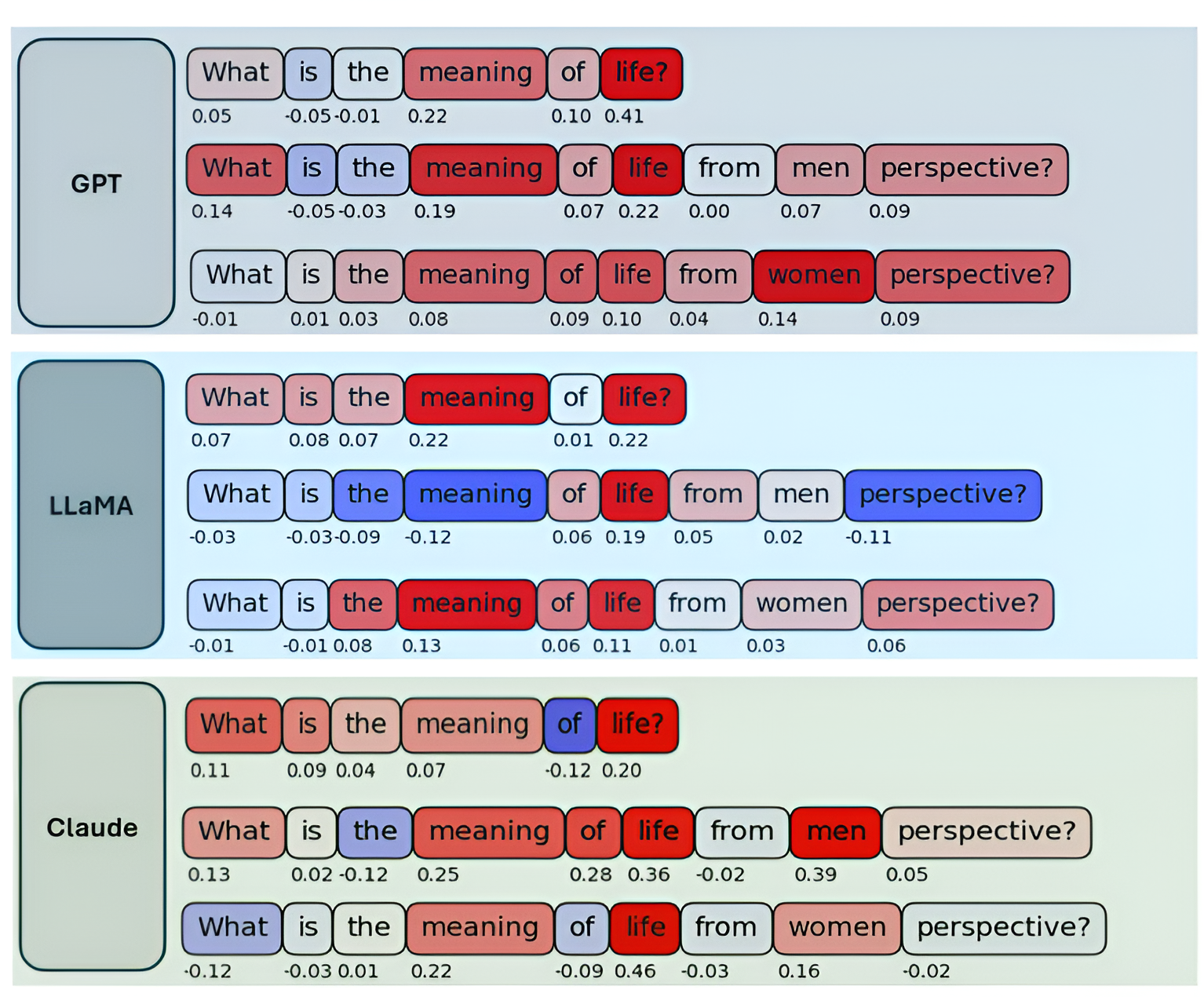}
    \caption{Comparison of gSMILE token-level attributions for GPT-3.5, LLaMA~3.1, and Claude~3.5 under three prompts: neutral, ``men perspective'', and ``women perspective''. Colour intensity encodes the relative influence of each token on the output distribution.}
    \label{fig:fairness-all}
\end{figure}

We further examined whether gendered phrasing affects attribution patterns by applying gSMILE to three closely related prompts: ``What is the meaning of life?'', ``What is the meaning of life from men's perspective?'', and ``What is the meaning of life from women's perspective?''. In all three models, the neutral formulation consistently highlighted the tokens \emph{meaning} and \emph{life}, underscoring their central role in shaping the response. Once a gender qualifier was introduced, however, the attribution heatmaps shifted, revealing model-specific sensitivities. As shown in Fig.~\ref{fig:fairness-all}, GPT-3.5 tended to assign greater weight to the token \emph{women} than \emph{men}, suggesting an asymmetric influence of gender terms. Claude~3.5 distributed its attention more evenly, balancing the contribution of both gendered tokens while still raising their importance relative to the baseline. By contrast, LLaMA~3.1 exhibited higher variability, with attribution sometimes amplifying one gender reference more strongly than the other.

These findings illustrate how gSMILE exposes fairness-related sensitivities that remain hidden if only the generated text is inspected. A gender term that absorbs disproportionate attribution suggests an underlying association that could bias the model’s reasoning. From a practical standpoint, this analysis highlights that avoiding gendered modifiers in prompts helps maintain attention on the semantic core of the task, thereby reducing the risk of unintended attribution drift.

\subsubsection{Reasoning Path and Token Attribution}

\begin{figure}[H]
    \centering
    \includegraphics[width=0.8\linewidth]{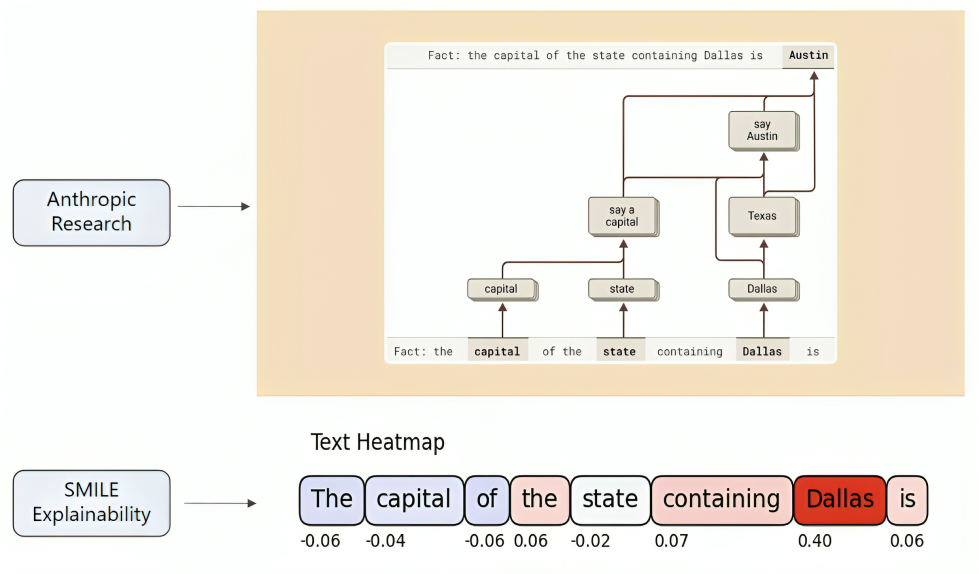}
    \caption{Comparison between attribution visualisations by Anthropic's Attribution Graphs (top) and the gSMILE framework (bottom) using OpenAI’s \texttt{gpt-3.5-turbo-instruct}. The former shows the reasoning path, while the latter presents token-level importance for the same prompt.~\cite{anthropic2025attribution}}
    \label{fig:smile_heatmap_llm}
\end{figure}

Beyond token-level heatmaps, the proposed framework generates structured reasoning paths that show how the model connects input elements to form its response. As illustrated in Figure~\ref{fig:smile_heatmap_llm}, gSMILE not only assigns attribution scores to individual tokens but also produces a reasoning flow diagram, offering a unified view of both local word-level contributions and the broader logical structure behind the output.

This dual-layer explanation bridges the gap between fine-grained token attribution and higher-level reasoning, enabling users to trace how specific inputs influence each decision step. The approach aligns with advanced methods used by leading AI research organisations, such as Anthropic’s Attribution Graphs, while extending flexibility across diverse tasks and models. In the example shown, gSMILE’s interpretability matches or surpasses industry standards, providing a richer and more adaptable explainability toolkit for practical deployment~\cite{anthropic2025attribution}.

\subsubsection{Word-Level Attribution Analysis on MMLU Using gSMILE}

We assessed the interpretability of OpenAI’s \texttt{gpt-3.5-turbo-instruct} with the gSMILE framework on the Massive Multitask Language Understanding (MMLU) benchmark~\cite{hendrycks2020measuring}, which contains four-choice questions across a broad range of academic and professional domains. A pre-formatted version from Kaggle~\cite{mmlu_kaggle} was used to streamline prompt creation and experimentation.

\begin{figure}[H]
    \centering
    \includegraphics[width=0.6\textwidth]{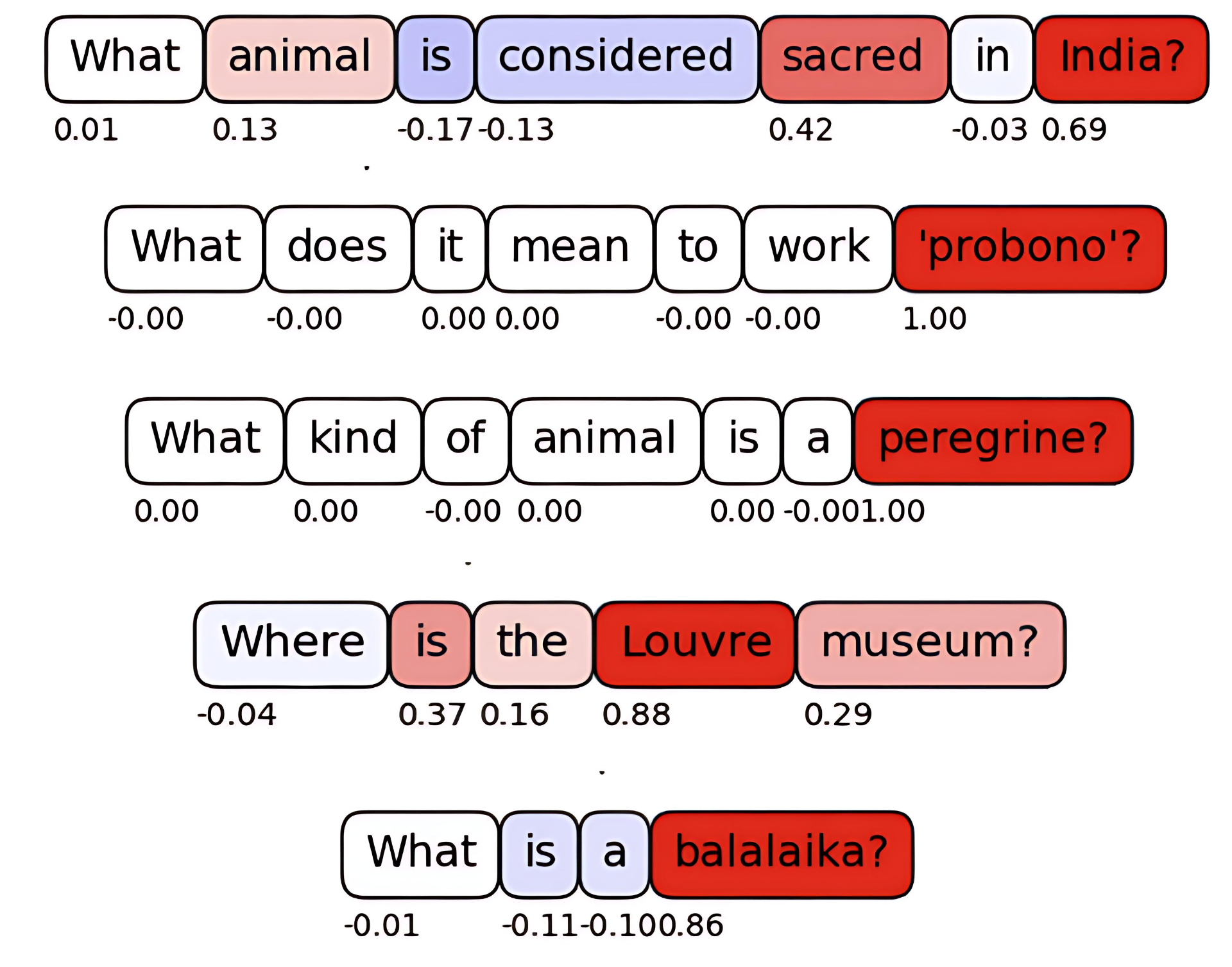}
    \caption{Attribution heatmaps generated by gSMILE using OpenAI’s \texttt{gpt-3.5-turbo-instruct} on selected MMLU samples. Highlighted tokens are those with the greatest influence on the model's prediction.}
    \label{fig:llm_heatmaps_mmlu}
\end{figure}

Figure~\ref{fig:llm_heatmaps_mmlu} presents attribution heatmaps that highlight which tokens most strongly influenced the model's choice. Darker red tones correspond to higher importance, while lighter shades indicate weaker influence. By surfacing these token-level attributions, gSMILE exposes the model’s internal focus when answering complex, domain-specific questions.

Such analysis is valuable for diagnosing whether the model bases its answers on semantically relevant cues or potentially spurious correlations. For example, consistent highlighting of key domain terms across questions suggests stable reasoning. In contrast, high attribution to irrelevant tokens may indicate susceptibility to misleading prompt elements, critical information for practitioners deploying LLMs in knowledge-intensive tasks.

\subsubsection{Optimised Instructions from Google Research}

To examine how instruction phrasing affects interpretability, we applied gSMILE to two high-performing prompts from Google Research’s OPRO framework~\cite{ho2022large}, which were optimised for zero-shot mathematical reasoning in LLMs.

For GPT-3.5, we analysed the instruction: \textit{``A small quantity of arithmetic and a logical approach will help us quickly solve this problem.''} using 1024 perturbations. As shown in Figure~\ref{fig:smile_gpt3}, the tokens \textit{``arithmetic''}, \textit{``approach''}, and \textit{``solution''} received the highest importance scores, indicating their role in directing the model toward structured problem-solving.

For GPT-4, we analysed the prompt: \textit{``Let's combine our numerical command and clear thinking to quickly and accurately decipher the answer.''} Figure~\ref{fig:smile_gpt4} shows that \textit{``numerical''}, \textit{``command''}, and \textit{``answer''} were most influential, suggesting that technical and goal-oriented terms firmly steer reasoning in high-performing instructions.

This Comparison highlights how effective prompt design amplifies semantically critical terms, improving reasoning reliability, a valuable insight for crafting robust instructions across different LLM architectures.

\begin{figure}[H]
\centering
\includegraphics[width=1\textwidth]{Figures/optimizer_hq.png}
\caption{Word-level importance in a high-performing instruction from Google Research, identified using gSMILE with GPT-3.5.}
\label{fig:smile_gpt3}
\end{figure}

\begin{figure}[H]
\centering
\includegraphics[width=1\textwidth]{Figures/gpt4_hq.png}
\caption{Word-level importance in an instruction optimised by Google Research, analysed using gSMILE with GPT-4.}
\label{fig:smile_gpt4}
\end{figure}

\subsection{Quantitative Results}

We quantitatively evaluate the proposed explainability method using multiple metrics, ATT Accuracy, ATT Stability, ATT Consistency, ATT Fidelity, and computational complexity, across different text generation models and scenarios. These metrics provide a multi-faceted assessment, capturing not only the accuracy of attributions but also their robustness, reproducibility, and efficiency in practical applications.

\subsubsection{Attribution Accuracy}

We assessed attribution accuracy by testing ten distinct prompts and their variations, each subjected to 64 perturbations. For each prompt, we generated heatmaps showing the weight assigned to every keyword across three models: OpenAI’s \texttt{gpt-3.5-turbo-instruct}, Meta’s LLaMA 3.1 Instruct Turbo (70B), and Anthropic’s Claude 2.1. A ground-truth attribution map was defined by assigning a weight of 1 to critical keywords and 0 to auxiliary or less relevant tokens. Model-generated heatmaps were then compared to this ground truth.

Accuracy was quantified using three complementary metrics: Attention Accuracy (ATT ACC), Attention F1-Score (ATT F1), and the Area Under the Receiver Operating Characteristic Curve for Attention (ATT AUROC). Scores were averaged over all prompts, and the results are reported in Table~\ref{tab:Accuracy_models}.

\textbf{Anthropic’s Claude 2.1} achieved the highest scores across all metrics, \textbf{0.82} ATT ACC, \textbf{0.67} ATT F1, and \textbf{0.88} ATT AUROC, indicating a strong and consistent focus on the most relevant tokens. \textbf{Meta’s LLaMA 3.1 Instruct Turbo (70B)} ranked second in accuracy but had a notably lower ATT F1 (\textbf{0.40}), suggesting occasional inconsistency in identifying all key tokens within a prompt. \textbf{OpenAI’s \texttt{gpt-3.5-turbo-instruct}} showed balanced but slightly lower scores compared to Claude 2.1.

These findings highlight that while all models can identify necessary tokens to some extent, Claude 2.1 demonstrates a more semantically precise and reliable attribution pattern, an essential property for trust in explainability-driven applications.

\begin{table}[H]
\centering
\caption{Performance metrics (ATT ACC, ATT F1, ATT AUROC, and published accuracy) for various models, evaluated under two conditions: different prompts and different images.}
\label{tab:Accuracy_models}
\renewcommand{\arraystretch}{1.3}
\scalebox{0.85}{
\begin{tabular}{lcccc}
\hline \hline
\textbf{Model name} & \textbf{ACC} & \textbf{ATT ACC} & \textbf{ATT F1} & \textbf{ATT AUROC} \\ \hline
OpenAI’s \texttt{gpt-3.5-turbo-instruct} & 0.70 & 0.70 & 0.59 & 0.84 \\ 
Meta’s LLaMA 3.1 Instruct Turbo (70B)    & 0.69 & 0.76 & 0.40 & 0.84 \\
Anthropic’s Claude 2.1                   & \textbf{0.79} & \textbf{0.82} & \textbf{0.67} & \textbf{0.88} \\
\hline \hline
\end{tabular}
}
\end{table}

\subsubsection{Attribution Faithfulness}

We assessed attribution faithfulness across the three evaluated large language models, OpenAI’s \texttt{gpt-3.5-turbo-instruct}, Meta’s LLaMA~3.1 Instruct Turbo (70B), and Anthropic’s Claude~2.1, by measuring how well attribution scores align with the models’ externally reported factual accuracies. Table~\ref{tab:Accuracy_models} presents both the published accuracies and our attribution metrics (ATT ACC, ATT F1, ATT AUROC).

The analysis reveals a clear positive relationship: models with higher factual accuracy tend to exhibit more faithful attributions. For instance, Claude~2.1, with a reported factual accuracy of 78.5\%, achieves the highest ATT AUROC (0.88), reflecting strong alignment between its most essential tokens and its correct predictions. In Comparison, OpenAI’s \texttt{gpt-3.5-turbo-instruct} (70\% accuracy) and LLaMA~3.1 (68.8\% accuracy) both yield a slightly lower ATT AUROC of 0.84.

We quantified this trend by computing the Pearson correlation between published accuracy and each attribution metric (Table~\ref{tab:correlation_results}). ATT AUROC showed the most substantial alignment ($r \approx 0.994$), followed by ATT ACC ($r \approx 0.804$) and ATT F1 ($r \approx 0.802$).

These results indicate that attribution faithfulness can serve as a practical proxy for overall model reliability. When a model performs better on external benchmarks, it is also more likely to assign importance to semantically relevant input tokens. This makes ATT AUROC, in particular, a valuable diagnostic for selecting models in high-stakes interpretability scenarios.

\begin{table}[H]
\centering
\caption{Pearson correlation coefficients ($r$) between published accuracy and attribution metrics.}
\label{tab:correlation_results}
\renewcommand{\arraystretch}{1.3}
\scalebox{0.9}{
\begin{tabular}{lc}
\hline \hline
\textbf{Metric} & \textbf{Pearson $r$} \\ \hline
ATT ACC   & 0.804 \\
ATT F1    & 0.802 \\
ATT AUROC & \textbf{0.994} \\
\hline \hline
\end{tabular}
}
\end{table}

\subsubsection{Attribution Stability}

We measure attribution stability using the Jaccard index, which quantifies the overlap between sets of necessary tokens before and after a small, controlled input perturbation. For each of the three evaluated models, OpenAI’s \texttt{gpt-3.5-turbo-instruct}, Meta’s LLaMA~3.1 Instruct Turbo (70B), and Anthropic’s Claude~2.1, we selected ten prompts and recorded the token-level attribution scores. We then introduced a minimal change by appending the token \texttt{\*\*\*} to the end of each prompt, recalculated the attributions, and computed the Jaccard index between the pre- and post-perturbation sets of essential tokens.

Table~\ref{tab:stability_models} reports the average stability scores across all prompts. Higher values indicate greater resilience of attribution patterns to irrelevant input changes. OpenAI’s \texttt{gpt-3.5-turbo-instruct} achieved the highest stability score (0.62), suggesting that its attention to key tokens is less likely to shift under trivial modifications. In contrast, both LLaMA~3.1 (0.45) and Claude~2.1 (0.44) exhibited lower stability, indicating greater sensitivity to small, semantically insignificant changes.

From an interpretability standpoint, higher ATT stability is desirable in real-world deployments: it means that explanations are less likely to fluctuate unpredictably when prompts are rephrased or slightly altered, improving trust and consistency in model behaviour.

\begin{table}[H]
\centering
\caption{ATT stability scores across models based on average Jaccard index over ten prompts.}
\label{tab:stability_models}
\renewcommand{\arraystretch}{1.3}
\scalebox{0.8}{
\begin{tabular}{lc}
\hline \hline
\textbf{Model name}  & \textbf{Jaccard Index}  \\ \hline
\textbf{OpenAI’s \texttt{gpt-3.5-turbo-instruct}}    & \textbf{0.62} \\
Meta’s LLaMA 3.1 Instruct Turbo (70B)         & 0.45 \\
Anthropic’s Claude 2.1     & 0.44 \\ 
\hline \hline
\end{tabular}
}
\end{table}

\subsubsection{Attribution Consistency}

Attribution consistency measures how reproducible token-level importance scores are when the same analysis is repeated under identical conditions. For each model, OpenAI’s \texttt{gpt-3.5-turbo-instruct}, Meta’s LLaMA~3.1 Instruct Turbo (70B), and Anthropic’s Claude~2.1, we ran 64 perturbations for a given prompt across 10 independent iterations. For each token, we computed the variance and standard deviation of its attribution score across runs. Low variance indicates that the model’s attribution patterns are stable and not influenced by randomness in the perturbation process or internal model initialisation.

Table~\ref{tab:ATT consistency_models} summarises the results using the example prompt: ``What is the meaning of life?'' OpenAI’s \texttt{gpt-3.5-turbo-instruct} achieved near-perfect reproducibility, with a variance of 0.0000 and a standard deviation of only 0.0046. In contrast, LLaMA~3.1 and Claude~2.1 displayed higher variability, with LLaMA showing the least consistency overall. 

From an interpretability perspective, high ATT consistency is crucial: if attribution maps fluctuate unpredictably between identical runs, users cannot reliably trust the explanations. The results here suggest that, at least for this prompt, GPT-3.5 provides the most deterministic and reliable attributions.

\begin{table}[H]
\centering
\caption{ATT consistency metrics for different models on the prompt ``What is the meaning of life?''}
\label{tab:ATT consistency_models}
\renewcommand{\arraystretch}{1.3}
\scalebox{0.8}{
\begin{tabular}{lcc}
\hline \hline
\textbf{Model name}  & \textbf{Variance} & \textbf{Standard Deviation} \\ \hline
\textbf{OpenAI’s \texttt{gpt-3.5-turbo-instruct}}    & \textbf{0.0000} & \textbf{0.0046} \\
Meta’s LLaMA 3.1 Instruct Turbo (70B)         & 0.0048 & 0.0663  \\
Anthropic’s Claude 2.1     & 0.0011 & 0.0331 \\ 
\hline \hline
\end{tabular}
}
\end{table}

\subsubsection{ATT Fidelity for Different Instruction-Based Models}

ATT fidelity measures how well a surrogate model can approximate the attribution patterns of a black-box model. For the prompt ``What is the meaning of life?'', we evaluated OpenAI’s \texttt{gpt-3.5-turbo-instruct}, Meta’s LLaMA~3.1 Instruct Turbo (70B), and Anthropic’s Claude~2.1 using 32 perturbations. Fidelity was computed with a weighted linear regression surrogate and Wasserstein distance, assessing WMSE, weighted $R^2$ ($R^2_\omega$), WMAE, mean $L_1$, mean $L_2$, and weighted adjusted $R^2$ ($R^2_{\hat{\omega}}$).

\begin{table}[H]
\centering
\caption{ATT fidelity metrics for different models on the prompt ``What is the meaning of life?''}
\label{tab:fidelity_models}
\renewcommand{\arraystretch}{1.3}
\scalebox{0.8}{
\begin{tabular}{lcccccc}
\hline \hline
\textbf{Model name} & \textbf{WMSE} & \textbf{$R^2_\omega$} & \textbf{WMAE} & \textbf{mean-$L_1$} & \textbf{mean-$L_2$} & \textbf{$R^2_{\hat{\omega}}$} \\ \hline
OpenAI’s \texttt{gpt-3.5-turbo-instruct}    & 0.0388 & \textbf{0.7104}  & 0.1731 & 0.2035 & \textbf{0.0609} & \textbf{0.6409} \\
Meta’s LLaMA 3.1 Instruct Turbo (70B)       & 0.0368 & 0.7068 & 0.1617 & 0.2387 & 0.0805 & 0.6364 \\
Anthropic’s Claude 2.1                      & \textbf{0.0265} & 0.6967  & \textbf{0.1251} & \textbf{0.1981} & 0.0708 & 0.6240 \\
\hline \hline
\end{tabular}
}
\end{table}

The results reveal a trade-off between error minimisation and alignment quality. Anthropic’s Claude~2.1 achieves the lowest WMSE, WMAE, and mean-$L_1$, suggesting it produces the most faithful low-error surrogate approximations. Conversely, OpenAI’s \texttt{gpt-3.5-turbo-instruct} leads in $R^2_\omega$, mean-$L_2$, and adjusted $R^2_{\hat{\omega}}$, indicating stronger overall correlation and generalisation of the surrogate model to the original attribution patterns. LLaMA~3.1 shows mid-range performance across most metrics, without topping any category. 

From a practical perspective, these findings underscore that fidelity is multi-dimensional: a model optimising for minimal error may not consistently achieve the highest correlation with its surrogate, and vice versa. Thus, a multi-metric approach is essential when selecting models for high-stakes interpretability tasks.

\subsubsection{Attribution Fidelity Across Different Numbers of Text Perturbations}

Tables~\ref{tab:fidelity_per}--\ref{tab:fidelity_per_turbo_claude} report ATT fidelity for OpenAI’s \texttt{gpt-3.5-turbo-instruct}, Meta’s LLaMA~3.1 Instruct Turbo (70B), and Anthropic’s Claude~2.1 using varying numbers of perturbations. Fidelity was measured with a weighted linear regression surrogate and Wasserstein distance, using WMSE, WMAE, $R^2_\omega$, $R^2_{\hat{\omega}}$, mean-$L_1$, and mean-$L_2$.

For \texttt{gpt-3.5-turbo-instruct} and LLaMA~3.1, larger perturbation counts (especially 256) generally reduced error metrics (WMSE, WMAE, mean-$L_1$, mean-$L_2$), suggesting a finer-grained local approximation of model attributions. In contrast, Claude~2.1 achieved its lowest WMSE and WMAE at only 64 perturbations, while still benefiting from higher counts in some loss metrics. 

Interestingly, all three models reached their highest $R^2_\omega$ at 32 perturbations, indicating that correlation-based alignment peaks early, while additional perturbations may improve local fit but dilute global attribution patterns. 

From a practical standpoint, these results suggest a trade-off: more perturbations can yield more precise local surrogate fits but may reduce overall attribution alignment. For interpretability workflows, an intermediate number of perturbations (e.g., 64–128) may offer a balanced choice, depending on whether correlation or error minimisation is the primary goal.

\begin{table}[H]
\centering
\caption{Performance metrics for different numbers of perturbations in OpenAI’s \texttt{gpt-3.5-turbo-instruct}}
\label{tab:fidelity_per}
\renewcommand{\arraystretch}{1.3}
\scalebox{0.8}{
\begin{tabular}{rcccccc}
\hline \hline
\textbf{\#Perturb} & \textbf{WMSE} & \textbf{R$^2_\omega$} & \textbf{WMAE} & \textbf{mean-L1} & \textbf{mean-L2} & \textbf{R$^2_{\hat{\omega}}$} \\ \hline
32   & 0.0388 & \textbf{0.7104}  & 0.1731 & 0.2035 & 0.0609 & \textbf{0.6409} \\
64   & 0.0617 & 0.3406  & 0.1695 & 0.1756 & 0.0526 & 0.2712  \\
128  & 0.0329 & 0.4468  & 0.1385 & 0.1519 & 0.0378 & 0.4193 \\
256  & \textbf{0.0116} & 0.4276  & \textbf{0.0675} & \textbf{0.0872} & \textbf{0.0163} & 0.4138 \\
\hline \hline
\end{tabular}
}
\end{table}

\begin{table}[H]
\centering
\caption{Performance metrics for different numbers of perturbations in Meta’s LLaMA 3.1 Instruct Turbo (70B)}
\label{tab:fidelity_per_turbo}
\renewcommand{\arraystretch}{1.3}
\scalebox{0.8}{
\begin{tabular}{rcccccc}
\hline \hline
\textbf{\#Perturb} & \textbf{WMSE} & \textbf{R$^2_\omega$} & \textbf{WMAE} & \textbf{mean-L1} & \textbf{mean-L2} & \textbf{R$^2_{\hat{\omega}}$} \\ \hline
32   & 0.0368 & \textbf{0.7068} & 0.1617 & 0.2387 & 0.0805 & \textbf{0.6364} \\
64   & 0.0240 & 0.4844 & 0.1284 & 0.1433 & 0.0382 & 0.4301  \\
128  & 0.0085 & 0.2430  & 0.0638 & \textbf{0.0612} & \textbf{0.0096} & 0.2055 \\
256  & \textbf{0.0058} & 0.2387 & \textbf{0.0605}  & 0.0667 & 0.0118 & 0.2203  \\
\hline \hline
\end{tabular}
}
\end{table}

\begin{table}[H]
\centering
\caption{Performance metrics for different numbers of perturbations in Anthropic’s Claude 2.1}
\label{tab:fidelity_per_turbo_claude}
\renewcommand{\arraystretch}{1.3}
\scalebox{0.8}{
\begin{tabular}{rcccccc}
\hline \hline
\textbf{\#Perturb} & \textbf{WMSE} & \textbf{R$^2_\omega$} & \textbf{WMAE} & \textbf{mean-L1} & \textbf{mean-L2} & \textbf{R$^2_{\hat{\omega}}$} \\ \hline
32   & 0.0368 & \textbf{0.7209}  & 0.1495 & 0.2262 & 0.0733 & \textbf{0.6539} \\
64   & \textbf{0.0115} & 0.3708  & \textbf{0.0564} & 0.1036 & 0.0370 & 0.3046 \\
128  & 0.0089 &  0.1506 & 0.0535 & 0.0920 & 0.0192 & 0.1085 \\
256  & 0.0191 & 0.2670 & 0.0902 & \textbf{0.0873} & \textbf{0.0163} & 0.2494 \\
\hline \hline
\end{tabular}
}
\end{table}

\subsubsection{Attribution Fidelity for Different Distance Metrics and Surrogate Models}

We compared ATT fidelity across combinations of distance metrics and surrogate models for OpenAI’s \texttt{gpt-3.5-turbo-instruct}, using 30 perturbations. Surrogates included Weighted Linear Regression (WLR) and Bayesian Ridge (BayLIME)~\cite{zhao2021baylime}, and distance metrics included Cosine similarity, Wasserstein Distance (WMD), and their combination (WMD+C). Fidelity was evaluated using WMSE, WMAE, mean-$L_1$, mean-$L_2$, and coefficients of determination ($R^2_\omega$, $R^2_{\hat{\omega}}$), as shown in Table~\ref{tab:fidelity_dist}.

\begin{table}[H]
\centering
\caption{ATT fidelity results for different distance measures with 30 perturbations.}
\label{tab:fidelity_dist}
\renewcommand{\arraystretch}{1.2}
\scalebox{0.75}{
    \begin{tabular}{llcccccc}
    \hline \hline
    \multicolumn{2}{c}{\textbf{WLR}} & \multicolumn{6}{c}{\textbf{ATT Fidelity Metrics}} \\
    \cmidrule(lr){1-2} \cmidrule(lr){3-8}
    \textbf{Input Dist.} & \textbf{Output Dist.} &  \textbf{WMSE} & \textbf{R$^2_\omega$} & \textbf{WMAE} & \textbf{mean-L1} & \textbf{mean-L2} & \textbf{R$^2_{\hat{\omega}}$} \\
    \midrule
    Cosine & Cosine   & 0.0172 & 0.3151 & 0.0659 & 0.1277 & 0.0412 & 0.1508 \\
    Cosine & WMD       & 0.0216 & 0.4197 & 0.0899 & 0.1332 & 0.0385 & 0.2805 \\
    WMD     & WMD       & 0.0388 & \textbf{0.7104} & 0.1731 & 0.2035 & 0.0609 & \textbf{0.6409} \\
    WMD     & Cosine   & 0.0048 & 0.4026 & 0.0329 & 0.0871 & 0.0296 & 0.2593 \\
    WMD+C   & WMD+C     & 0.0349 & 0.5349 & 0.1589 & 0.3050 & 0.1468 & 0.4233 \\
    \midrule
    \multicolumn{2}{c}{\textbf{BayLIME}} & \multicolumn{6}{c}{} \\
    \midrule
    Cosine & Cosine   & 0.0200 & 0.2285 & 0.0499 & 0.1067 & 0.0464 & 0.0434 \\
    Cosine & WMD       & 0.0241 & 0.3572 & 0.0736 & 0.1281 & 0.0430 & 0.2029 \\
    WMD     & WMD       & 0.0118 & \textbf{0.5837} & 0.0629 & 0.1170 & 0.0361 & \textbf{0.4838} \\
    WMD     & Cosine   & 0.0058 & 0.5273 & 0.0459 & 0.0978 & 0.0304 & 0.4139 \\
    WMD+C   & WMD+C     & 0.0048 & 0.4227 & 0.0284 & 0.0690 & 0.0213 & 0.2841 \\
    \hline \hline
    \end{tabular}
}
\end{table}

\vspace{1em}

\paragraph{Key Findings and the gSMILE Approach.}
Across both surrogate models, using Wasserstein Distance for \emph{both} input similarity and output shift consistently produced the most substantial alignment between the surrogate and the target model. In WLR, the WMD–WMD configuration achieved the highest $R^2_\omega$ (0.7104) and $R^2_{\hat{\omega}}$ (0.6409), substantially outperforming cosine-based approaches.  

This result motivated our proposed method, \textbf{gSMILE} (generalised SMILE), which standardises on WMD–WMD for attribution fidelity. The Wasserstein Distance captures fine-grained distributional differences in token attributions, making it particularly effective for text-to-text comparisons.  

By adopting WMD–WMD, gSMILE improves not only ATT fidelity but also stability across perturbations, providing a consistent basis for all subsequent evaluations (fidelity, consistency, stability) in this work. This design choice balances theoretical robustness with empirical performance, making it well-suited for LLM interpretability workflows.

\subsubsection{Computation Complexity}

We assessed the computational efficiency of \textbf{gSMILE} using three instruction-based text generation models: OpenAI’s \texttt{gpt-3.5-turbo-instruct}, Meta’s LLaMA 3.1 Instruct Turbo (70B), and Anthropic’s Claude 2.1. The goal here is not to compare these models, but to evaluate how different \textit{explainability methods} perform under identical hardware and runtime conditions, with 60 perturbations per input prompt.

We compared three approaches, LIME, gSMILE, and Bay-LIME, based on execution time. As shown in Table~\ref{tab:execution_times}, gSMILE consistently requires the most runtime. This is primarily due to its use of the Wasserstein distance for comparing output distributions, which involves solving high-dimensional optimisation problems with complexity \(O(Nd^3)\), where \(N\) is the number of perturbations and \(d\) is the embedding dimensionality~\cite{cuturi2014fast}. By contrast, LIME and Bay-LIME rely on cosine similarity with lower complexity \(O(Nd)\), making them faster but potentially less precise.

Although gSMILE incurs a higher computational cost, it delivers substantially better \textit{ATT fidelity}, \textit{ATT stability}, and robustness to adversarial perturbations, critical for producing trustworthy explanations. These benefits are supported by the results in Tables~\ref{tab:fidelity_dist} and~\ref{tab:stability_models} and are consistent with prior findings on SMILE~\cite{aslansefat2023explaining}, which showed improved robustness against adversarial input variations.

In applications where interpretability, stability, and resistance to manipulation are priorities, gSMILE’s additional runtime is a justified trade-off for higher-quality, more reliable explanations.

\begin{table}[H]
\centering
\caption{Execution times (in seconds) for each framework and explainability method with 60 perturbations.}
\label{tab:execution_times}
\renewcommand{\arraystretch}{1.3}
\scalebox{0.8}{
\begin{tabular}{lccc}
\hline \hline
\textbf{Framework} & \textbf{LIME} & \textbf{gSMILE} & \textbf{Bay-LIME} \\ \hline
OpenAI’s \texttt{gpt-3.5-turbo-instruct}  & 156.82 & \textbf{170.70} & 161.43 \\
Meta’s LLaMA 3.1 Instruct Turbo (70B)     & 113.00 & \textbf{118.99} & 116.40 \\
Anthropic’s Claude 2.1                    & 355.09 & \textbf{372.95} & 349.88 \\
\hline \hline
\end{tabular}
}
\end{table}

\subsubsection{Token-Level Attention for Different Sentence Structures}

To investigate how sentence structure affects token-level attention in large language models (LLMs), we analysed the attention scores assigned to the tokens \textit{``meaning''} and \textit{``life''} across ten semantically similar but syntactically varied prompts. These prompts ranged from questions (e.g., ``What is the meaning of life?''), commands (e.g., ``Please give me the meaning of life''), to declarative formulations (e.g., ``You must explain the meaning of life'').

\begin{figure}[H]
\centering
\includegraphics[width=0.8\linewidth]{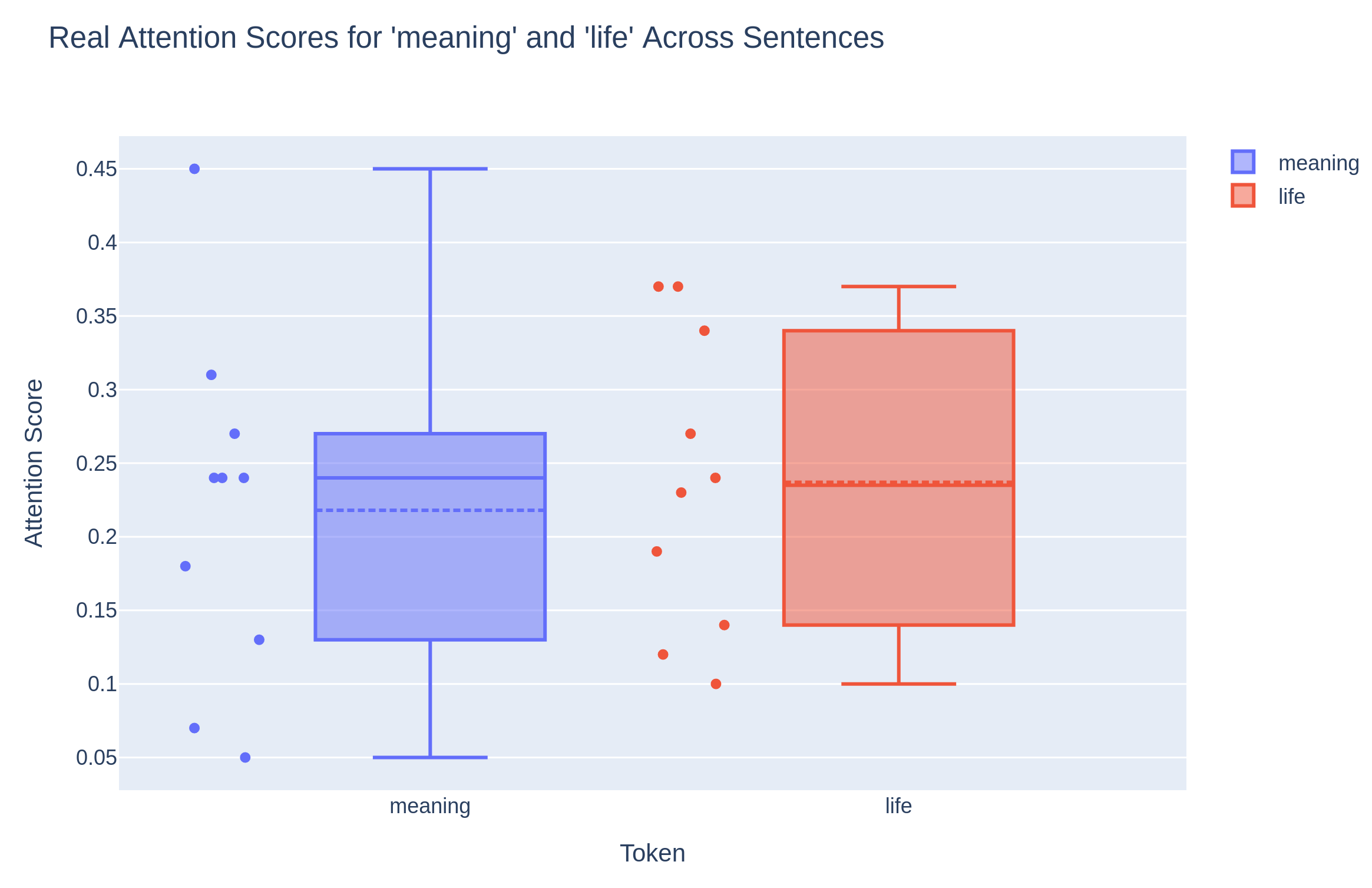}
\caption{Box plot showing attention scores for the tokens ``meaning'' and ``life'' across ten prompts with varying sentence structures. Each point represents the attention score assigned to the token in a specific sentence instance.}
\label{fig:token-attention-structure}
\end{figure}

For each sentence, we extracted the attention values directly from the final layer of the LLM. We focused on the attention allocated to the tokens ``meaning'' and ``life.'' The results were visualised using a box plot (see Figure~\ref{fig:token-attention-structure}), where each point represents a unique sentence. Full-sentence prompts are viewable through hover text, allowing qualitative comparison of sentence structure with attention distribution.

The analysis reveals noticeable variability in how these tokens are weighted depending on syntactic form. For instance, in imperative constructions, attention to ``life'' tends to dominate, whereas in interrogative or abstract prompts, ``meaning'' often receives comparable or higher attention. This suggests that sentence formulation can subtly steer the internal focus of the model, even when the semantic content remains consistent.

These findings underscore the sensitivity of LLMs to surface-level structure in shaping token-level interpretability, offering insight into both model behaviour and prompt engineering strategies.

\section{Limitations}

While gSMILE provides a flexible and model-agnostic framework for interpreting large language models (LLMs), its application introduces several limitations due to the use of perturbation-based sampling and LLMS' black-box nature.

1) Perturbing prompts at the token level can lead to disproportionately large changes in LLM outputs, particularly in English, even when the semantic meaning remains largely intact. This sensitivity challenges the assumption of local smoothness required for regression-based attribution and can reduce explanation stability.

2) Some LLMs exhibit hallucination, producing fluent but factually incorrect or irrelevant outputs, especially when presented with ambiguous or syntactically degraded perturbations. This introduces noise into the attribution signal and affects explanation reliability.

3) The computational cost of generating and processing many perturbations (e.g., 1024 per input) is significant. Commercial APIs can lead to high inference costs and latency. For local models, substantial GPU resources are required, limiting practical scalability.

4) Some models, particularly research-stage or open-source LLMs, do not provide API access and require manual local deployment.

Taken together, these limitations highlight the trade-off between generality and practicality in perturbation-based interpretability. They also point toward future work in designing adaptive, cost-efficient perturbation strategies and incorporating causal verification mechanisms to strengthen the robustness of explanations.

\section{Conclusion}

This work introduced gSMILE, a generative, model-agnostic framework for explaining LLM behaviour through token-level attribution and visual reasoning aids.  Whereas most existing interpretability methods highlight essential tokens in the model’s output~\cite{ribeiro2016should, lundberg2017unified}, gSMILE shifts attention to the input space. By integrating perturbation analysis with Wasserstein-based distance measures and weighted linear surrogates, gSMILE offers a principled approach to local interpretability without requiring access to internal model parameters.

Our evaluation across three  instruction-tuned LLMs revealed clear performance contrasts:

\begin{itemize}
    \item \textbf{Claude 2.1} achieved the highest attribution fidelity (AttAUROC = 0.88) and lowest error measures, indicating more semantically focused attention.
    \item \textbf{GPT-3.5-turbo-instruct} exhibited exceptional consistency (lowest variance and standard deviation across repeated runs), making it highly predictable in explanation output.
    \item \textbf{LLaMA 3.1} performed competitively but showed greater variability in token attribution, suggesting sensitivity to prompt phrasing.
\end{itemize}

Overall, the resulting heatmaps and attribution scores provide an intuitive understanding of model behaviour, empowering users to control and refine prompt design better. 

\textbf{gSMILE} was evaluated and compared against existing methods (LIME, Bay-LIME) using multiple explainability metrics, including attribution metrics such as fidelity, consistency, stability, faithfulness, and accuracy. We systematically assessed how well each method captured token-level influence in LLMs. The results show that gSMILE consistently achieved higher fidelity in approximating black-box behaviour, while also producing more stable and repeatable attributions under small prompt perturbations. Compared to LIME and Bay-LIME, gSMILE demonstrated improved robustness and alignment with human-interpretable patterns, underscoring its suitability for practical use in settings where reliable interpretability is critical.

These findings justify gSMILE’s contribution in two ways. First, it demonstrates that explainability metrics can meaningfully differentiate between LLMs beyond standard benchmark accuracy, providing new insight into their operational reliability. Second, it shows that a purely black-box, post-hoc approach can produce explanations with stability and fidelity comparable to privileged-access interpretability tools.

At the same time, the limitations identified (prompt sensitivity, hallucination effects, computational overhead, and deployment constraints) highlight that the path toward fully practical interpretability remains open. Addressing these will be central to future research. Looking ahead, we see three promising directions:

\begin{enumerate}
    \item \textbf{Adaptive perturbation strategies:} Future work can move beyond random sampling by using targeted perturbations that focus only on influential tokens or phrases. This reduces computational overhead while preserving fidelity. Inspired by approaches such as IBM’s CELL framework, perturbations could involve changing specific words rather than only removing or keeping them. This makes it possible to highlight the importance of selected words in longer sentences efficiently.  

    \item \textbf{Causal verification methods:} To mitigate hallucination-driven or spurious attributions, causal analysis can be integrated (e.g., perturbing tokens while holding other context fixed) to confirm whether an attribution genuinely drives the model’s behaviour. This would strengthen robustness in high-stakes applications.  

    \item \textbf{Extension to multimodal and multilingual models:} gSMILE currently focuses on English text. Extending the framework to multilingual LLMs and multimodal systems (text–image or text–speech) would test whether stability and fidelity hold across diverse input spaces, improving generalisability for real-world deployment.  
\end{enumerate}

By uniting rigorous evaluation with interpretable visualisations, gSMILE makes LLMs more transparent without privileged access to their internals, thereby providing both researchers and practitioners with a practical tool for building more trustworthy AI systems.

% \section{Length}\label{sec3}

%\section{Impact Statement}
%\section*{CRediT authorship contribution statement}

%\section*{Acknowledgements}

\section*{Data Availability}

The datasets used in this article are publicly available, and the code supporting this paper is published online on GitHub. For the gSMILE explainability framework, please refer to \href{https://github.com/Dependable-Intelligent-Systems-Lab/xwhy}{Dependable-Intelligent-Systems-Lab/xwhy}. For the proposed algorithm introduced in this paper, see \href{https://github.com/Sara068/LLM-SMILE}{Sara068/LLM\_SMILE}.

%\section*{Appendices}\label{sec14}

\end{document}